\documentclass{ecai} 

\usepackage{latexsym}
\usepackage{amssymb}
\usepackage{amsmath}
\usepackage{amsthm}
\usepackage{booktabs}
\usepackage{enumitem}
\usepackage{graphicx}
\usepackage{color}
\usepackage{algorithm}
\usepackage{algpseudocode}
\usepackage[utf8]{inputenc}
\usepackage[T1]{fontenc}

\newcommand{\BibTeX}{B\kern-.05em{\sc i\kern-.025em b}\kern-.08em\TeX}

\begin{document}


\begin{frontmatter}


\paperid{2325} 


\title{An Arbitration Control for an Ensemble of Diversified DQN variants
in Continual Reinforcement Learning}


\author[A]{\fnms{Wonseo}~\snm{Jang}\orcid{0009-0004-2534-0578}}
\author[A,B]{\fnms{Dongjae}~\snm{Kim}\orcid{0000-0002-4513-9087}\thanks{Corresponding Author. Email: dongjaekim@dankook.ac.kr.}}

\address[A]{Department of AI-based Convergence, Dankook University}
\address[B]{Department of Artificial Intelligence, Dankook University}


\begin{abstract}
Deep reinforcement learning (RL) models, despite their efficiency in learning an optimal policy in static environments, easily loses previously learned knowledge (i.e., catastrophic forgetting). It leads RL models to poor performance in continual reinforcement learning (CRL) scenarios. To address this, we present an arbitration control mechanism over an ensemble of RL agents. 
It is motivated by and closely aligned with how humans make decisions in a CRL context using an arbitration control of multiple RL agents in parallel as observed in the prefrontal cortex.
We integrated two key ideas into our model: (1) an ensemble of RLs (i.e., DQN variants) explicitly trained to have diverse value functions and (2) an arbitration control that prioritizes agents with higher reliability (i.e., less error) in recent trials.
We propose a framework for CRL, an \emph{A}rbitration \emph{C}ontrol for an \emph{E}nsemble of \emph{D}iversified \emph{DQN} variants (\emph{ACED-DQN}). We demonstrate significant performance improvements in both static and continual environments, supported by empirical evidence showing the effectiveness of arbitration control over diversified DQNs during training. In this work, we introduced a framework that enables RL agents to continuously learn, with inspiration from the human brain.
\end{abstract}

\end{frontmatter}


\section{Introduction}

Continual learning in diverse contexts is a natural built-in skill for humans but rather challenging for reinforcement learning (RL) models. When learning something new, RL models tend to forget previously acquired experiences along with the policies, i.e., catastrophic forgetting~\citep{Abel2023,vandeVen2024}, resulting in poor performance in continual learning scenarios. Nonetheless, there are hurdles when designing RL models that perform well in continual reinforcement learning (CRL) environments lies in addressing the stability-plasticity dilemma, a balance between the ability to adapt to new information (plasticity) and the ability to retain previously acquired experiences (stability)~\citep{Khetarpal2020}. 

\begin{figure}[t]
\includegraphics[width=1\linewidth]{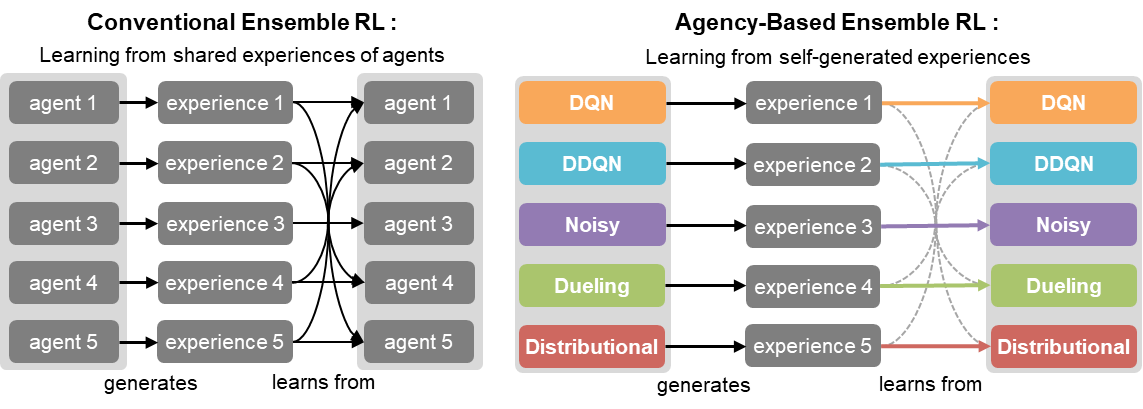}
\vspace{-0.56cm}
\caption{\textbf{A schematic view of conventional ensemble RL and the proposed, agency-based ensemble RL.} In agency-based ensemble RL with DQN variants, each agents most likely learns from self-generated experience, following the sense of agency.
}
\label{fig:intro}
\vspace{-0.3cm}
\end{figure}

Among the various remedies proposed in computer science to solve such continual learning problems, ensemble‐based approaches have emerged as a promising solution (Figure \ref{fig:intro}). By maintaining a set of diverse policies -- or distinct value-functions -- that are selectively activated or weighted, ensembles preserve adaptability to novel data, while retaining previously learned knowledge. In CRL specifically, techniques such as bootstrapped DQN~\citep{osband2016deep} have been proposed. It mitigates interference between tasks and improves the overall stability-plasticity trade-offs by training multiple value functions with different bootstrap samples. 

However, this approach has been shown to degrade performance in certain scenarios~\citep{Lin2024}. \citeauthor{Lin2024} demonstrated a counterintuitive phenomenon called \emph{curse of diversity}, which arises from the low proportion of self-generated data in the shared training set for each agent in ensemble, as well as from the difficulty individual members face in effectively learning from such highly off-policy data (Figure \ref{fig:intro}).

In contrast, humans have an outstanding ability to continually learn without suffering from \emph{curse of diversity}, despite the involvement of multiple valuation systems in the brain. In the frontal cortex, valuation process during decision-making are known to operate either in parallel or sequentially, depending on the goal of the tasks~\citep{Rushworth2012}. Indeed, multiple RL (e.g., model-based and model-free RL) systems are known to coexist in the human brain~\citep{Daw2011, glascher2012lesion}, along with an arbitration mechanism that dynamically allocates control between them in response to environmental changes~\citep{WanLee2014}. This has further been identified as a prefrontal mechanism to cope with increasing task complexity~\citep{kim2019} and to reduce errors beyond the conventional bias–variance tradeoff~\citep{kim2021}.

In this work, inspired by the ideas of diversified valuation mechanisms, we propose an Arbitration Control for an Ensemble of Diversified DQN variants (ACED-DQN), an approach that employs an ensemble of DQN variants (i.e., DQN~\citep{Mnih2013}, Double DQN~\citep{VanHasselt2015}, Noisy DQN~\citep{Fortunato2017}, Dueling DQN~\citep{Wang2015}, and Distributional DQN~\citep{Bellemare2017}) each trained with its own distinct value function. We introduce an arbitration control mechanism to dynamically select the most reliable agent -- that generates the least prediction errors~\citep{WanLee2014} -- among the DQN variants based on the current environmental context. 
Our method also addresses the curse of diversity~\citep{Lin2024}. Unlike conventional ensembles that share experiences, our approach allows each agent to learn primarily from its own generated experiences, avoiding performance degradation caused by highly off-policy data (Figure \ref{fig:intro}).

To summarize, our main contributions are:

\vspace{-0.3cm}

\begin{itemize}
\item We introduce an arbitration control mechanism for ensemble agents that assigns a greater weight to the most reliable RL agent based on its error.
\item We propose an agency-based experience sampling method that selectively sample experiences each agent has generated.
\item We present a novel approach, ACED-DQN, which demonstrates strong performance in CRL environments. Ablation studies indicate that arbitration control is the key contributing factor to its success.
\end{itemize}


\section{Related works}

\subsection{DQN variants}
Since its success in video games~\citep{Mnih2013}, DQN has become a widely adopted deep RL algorithm to solve many challenging tasks. It extends the Q-learning algorithm~\citep{Sutton2005} with the neural network-based value function approximation. However,  limitations such as low variance due to overestimation leave room for improvement. For such improvements, in addition to the vanilla {\bf DQN}, there are notable DQN variants that are used in our work. 

{\bf Double DQN} addresses the inherent exploitation of standard DQN by decoupling action selection from behavioral value evaluation in target valuation~\citep{VanHasselt2015}. It resolves the low-variance problem to some extent as its off-policy characteristics. 

{\bf Dueling DQN} separates value estimation in a neural network into two distinct heads: one estimating the state value, $V(s)$ and the other estimating the advantage, $A(s,a)=Q(s,a)-V(s)$, while sharing a convolutional encoder~\citep{Wang2015}. This improves generalization across actions that have similar values.

{\bf Noisy DQN} introduces learnable, parameterized noise into the network weights to replace traditional $\epsilon$-greedy exploration since many actions must be executed to collect the first reward~\citep{Fortunato2017}. This results in state-dependent exploration and more efficient behavior policy. 

{\bf Distributional DQN} (also known as \emph{C51}) models can approximate the return distribution via a fixed-support categorical distribution~\citep{Bellemare2017}. This enables the agent to represent uncertainty in value estimation, particularly beneficial in sparse-reward environments with high reward variability.

{\bf Rainbow} integrates aforementioned models into a single integrated agent~\citep{Hessel2017}. It combines: (1) Double DQN to mitigate overestimation bias; (2) Dueling DQN  to improve representation learning; (3) Noisy DQN to enable efficient exploration through parameterized noise in the network weights; and (4) Distributional DQN to model the full distribution of returns. This integrated agent improves learning efficiency and performance on discrete control environments, compared to when these components are used alone.

We provide implementation details and further comparisons of these models in the Supplementary.

\subsection{Ensemble RL}
The idea of ensemble RL to reduce variance and improve the robustness of the policy has been used in many deep RL algorithms~\citep{Januszewski2021,Osband2016,Schmitt2019}. Most of these works focus on homogeneous agents (i.e., agents that share the same learning algorithm), with some changes to encourage exploration or to produce robust value estimations for current tasks~\citep{Agarwal2019,Ishfaq2021,Openai2017,Peer2021}. {\bf Averaged DQN}~\citep{Ansehel2016} trains multiple Q-networks in parallel with different weight initializations and averages their outputs to reduce variance, resulting in a more stable policy for action selection.  {\bf Bootstrapped DQN}~\citep{Osband2016}, on the other hand, promotes deep exploration by maintaining multiple Q-heads trained on bootstrapped subsets of experience; at the beginning of each episode, one head is selected (greedily) for action selection. While these algorithms differ in their underlying mechanisms, both can be prone to divergence if one of the Q-networks becomes overly biased~\citep{Saphal2020}.  {\bf SUNRISE}~\citep{Lee2021} addresses the instability issues --mainly caused by bias-- using an ensemble-based weighted Bellman update, which constrains the diversity of the agents. Moreover, selecting actions using the Upper Confidence Bound (UCB) across the heads to encourage exploration. 

Although earlier ensemble RL methods have shown empirical success, their effectiveness has been largely limited to static environments. These approaches struggle in CRL settings, where task distributions change over time and individual agents tend to overfit to specific tasks~\citep{Khetarpal2020}. Recent studies have attempted to address this limitation by adaptively weighting each ensemble member based on their past performance~\citep{Lin2024, Agarwal2019}; however, it can suffer performance degradation due to the excessive diversity among agents.

\subsection{Curse of diversity}
Excessive diversity within ensembles can lead to degradation in performance~\citep{Lin2024}. This is a counterintuitive finding, especially considering the ensemble-based approaches are frequently used in continual learning in many machine learning problems. However, this becomes problematic in ensemble RL due to different dynamics of learning. 

When agents learn from a shared buffer but experience highly off-policy transitions generated by other ensemble members, they tend to converge to different estimates of the value function that are mutually inconsistent. For example, in standard prioritized experience replay (PER)~\citep{Schaul2015}, experiences are sampled based on their error magnitude, which serves as a prioritization metric reflecting the perceived learning utility. However, the priority does not account for the agent that originally generated the transition. This becomes problematic in ensemble settings, where agents with diverse value functions may evaluate and learn from each other's off-policy experiences, potentially leading to instability in value estimation.

These observations highlight the necessity of an alternative approach for experience replay of an ensemble RL: replaying experiences for an agent based on a sense of agency. In cognitive science, this is referred to ``sense of agency", that the subjective awareness of initiating and controlling one's actions and their consequences~\citep{moore2016sense}. Translating this to ensemble RL, it indicates attributing experiences to the originating agent (Figure \ref{fig:intro}; right panel). 


\section{Method}

\subsection{Continual Atari environments}
\begin{figure}[t]
\centering
\includegraphics[width=0.8\linewidth]{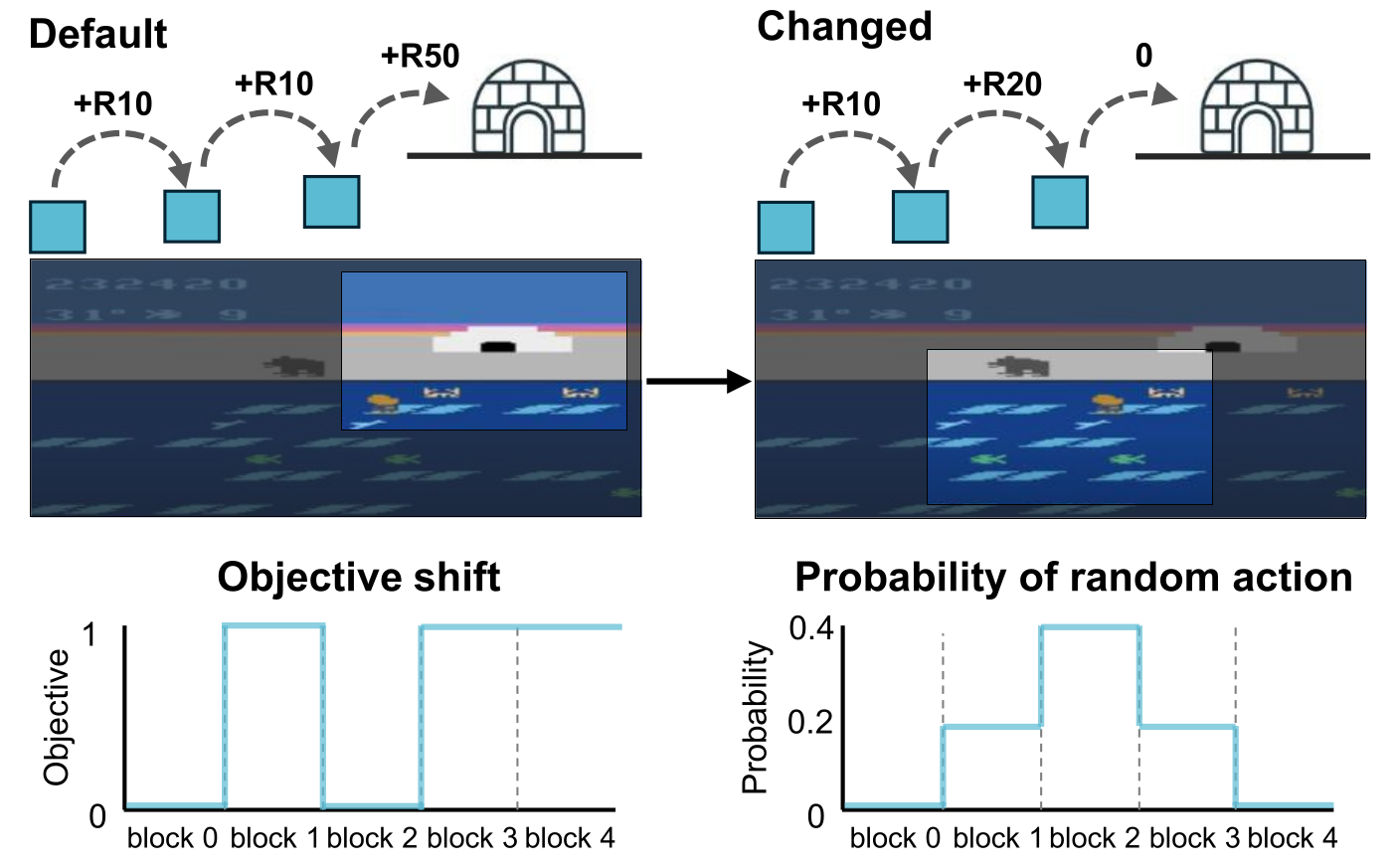}
\caption{{\bf Block conditions of continual Atari task.} 
Two environmental contexts are modified from block to block. There are two kinds of objective shifts: original (0) and modified (1). For the probability of random action, there are three levels of randomness (0, 0.2 and 0.4). 
}
\label{fig:2}
\end{figure}

To create environments for CRL, we adapted 26 Atari games from the Arcade Learning Environment~\citep{Ostrovski2017}~(Figure \ref{fig:2}). Atari games are well-established benchmark for evaluating RL agents in discrete action spaces. We specifically selected 26 games that are used in the Atari 100K benchmark, which measures the sample efficiency of RL agents~\citep{ye2021mastering}. In continual setting, each Atari game continues to 1 million steps, and changing the environmental contexts (objective and probability of random action) every 200K steps, resulting in five blocks of environmental contexts (Figure \ref{fig:2}). It is notable that although we chose a block length of 200K steps (rather than 100K) because of continual setting made baseline agents (i.e., Rainbow and SUNRISE) struggles, our findings are not limited to the specific length of each condition.

Changes in the environmental contexts, objective shift and probability of random action, are designed to change the learned knowledge. First, the objective shift is designed to completely change the reward function in a different direction. For example, in \emph{Frostbite}, the original objective (when objective is 0 in Figure \ref{fig:2}) is to survive by jumping over the ice plates until an igloo is built, and entering the igloo gives huge points. The modified objective (when the objective is 1 in Figure \ref{fig:2}) is rather jumping over the plates longer to survive, ignoring the points of entry into the igloo. 

Second, the probabiltiy of random action refers to the probability executing a random action instead of the intended one, which encourages the agent to adopt a more risk-averse policy. For example, in \emph{Space Invaders}—where the goal is to shoot down aliens—an agent might need to fire slightly earlier than usual, anticipating the chance that its intended action (e.g., firing) might not be executed due to stochastic interference. In our setup, we applied randomness independently to movement direction and firing actions when applicable. 

These environmental changes are designed to test RL models ability to adapt the variety of continual environmental changes. We fixed this block design (as in Figure \ref{fig:2}) for all Atari environments for fair comparison. Note that block 1 and block 3 are identical (both have a modified objective with small probability (0.2) of random action), which is intentionally designed to test if the continually learned knowledge maintained after other blocks (block 2). For more details such as the list of objectives of 26 games, refer to the Supplementary.

\subsection{ACED-DQN}
Here, we introduce ACED-DQN: {\bf A}rbitration {\bf C}ontrol for {\bf E}nsemble of {\bf D}iversified {\bf DQN} variants, which employs ensemble of DQN variants with arbitration control for selecting action based on agency of sampling experiences. Arbitration control relies on agents' reliability--which is a quantity how reliable the agent is in current environmental context-- for each agent. To calculate this, we used loss trends in DQN variants (i.e., mean-squared error; MSE). The reliability of agents are used to weight the Q-values of agents in action selection and experience sampling, thereby ensuring that the decision process is informed by agents currently best suited to the environment, while preserving a sense of agency.

\begin{figure}[t]
\includegraphics[width=1\linewidth]{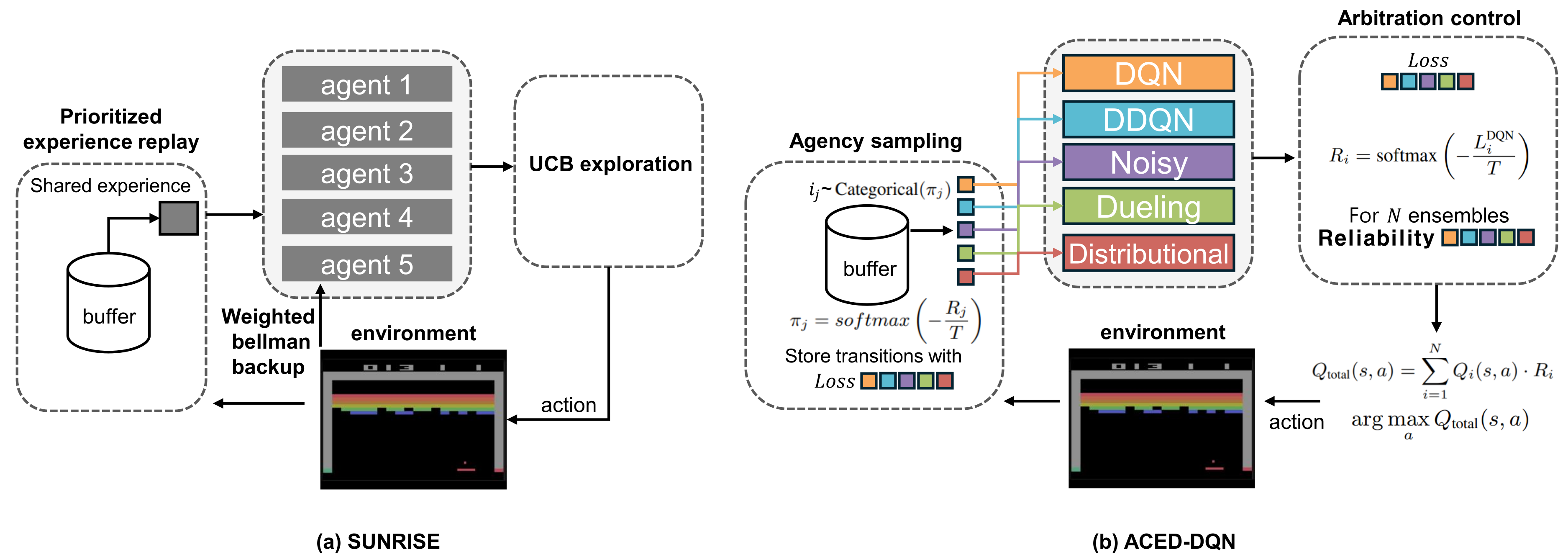}
\vspace{-0.5cm}
\caption{\textbf{Comparison between SUNRISE and ACED-DQN.} (a) Illustration of  SUNRISE with \(N\) number of homogeneous agents with one replay buffer. (b) Illustration of our framework. We consider an arbitration control over DQN variants.}
\label{fig:aced}
\end{figure}
\subsubsection{Action selection based on the arbitration control}

\begin{algorithm}[b]
\caption{Action Selection (Arbitration Control)}
\label{alg:1}
\begin{algorithmic}[1]
    \Statex \textbf{Input:} Q-values $\{Q_i(s,a)\}_{i=1}^N$, losses $\{L_i\}_{i=1}^N$
    \For{each agent $i = 1$ to $N$}
        \State Compute reliability using Eq.\ref{eq:Reliability}
    \EndFor
    \State Update reliability with momentum using Eq.\ref{eq:momentum}
    \State Clamp $\tilde{R}_i^{(t)}$ to $[R_{\min}, R_{\max}]$
    \State Normalize: $w_i \leftarrow \frac{\tilde{R}_i^{(t)}}{\sum_{j=1}^N \tilde{R}_j^{(t)}}$
    \State Aggregate Q-values: $\textstyle Q_{\text{total}}(s,a) = \sum_{i=1}^N w_i \cdot Q_i(s,a)$
    \State \textbf{Output:} $\textstyle a^* = \arg\max_a Q_{\text{total}}(s,a)$
\end{algorithmic}
\end{algorithm}

To train an ensemble, we consider an ensemble of $N$ diverse DQN variants where each agent's Q-value \( \{Q_i(s, a)\}_{i=1:N} \) is trained according to its respective learning method. We used vanilla DQN, Double DQN, Dueling DQN, Noisy DQN, and Distributional DQN(C51) as the agents of an ensemble. Since these RL models uses temporal difference (TD) learning-like updates, we can use MSE as signal for how unreliable the agent is in the current context. This approach is motivated by evidence that the human brain uses arbitration control guided by reliability signals derived from TD errors~\cite{WanLee2014}. Similarly, we propose an arbitration control mechanism that uses the MSE to find the most reliable agent in the current context (Algorithm \ref{alg:1}). 

To adaptively focus on reliable agents, we compute weights of each agent \(i\)-th Q-value based on reliability $R_i$ at time $t$ as follows:
\begin{equation}\label{eq:Reliability}
R_i^{(t)}= \text{softmax}\left(-\frac{L_i^{(t)}}{T}\right), \quad \text{for } i = 1, \dots, N,
\end{equation}
where  \( L_i \) is the MSE of agent \( i \), and  \( T \) is the temperature parameter. To stabilize the reliability estimation of each agent and reduce the variance caused by short-term fluctuations in loss error, we use the exponential moving average \(\tilde{R}_i^{(t)}\), updated at each time step \(t\) as:
\begin{equation}\label{eq:momentum}
\tilde{R}_i^{(t)} = (1 - \gamma) \cdot \tilde{R}_i^{(t-1)} + \gamma \cdot R_i^{(t)},
\end{equation}
where \( \gamma \) represents the smoothing factor that controls the contribution of the newly observed reliability. To prevent reliability values from converging to extreme values (i.e., near 0 or 1), we restrict  \(\tilde{R}_i^{(t)}\) to lie within a predefined range, using lower and upper bounds \(R_{\min}\) and \(R_{\max}\).
The reliability is then normalized to ensure proper weighting across the ensemble: $w_i \leftarrow \frac{\tilde{R}_i^{(t)}}{\sum_{j=1}^N \tilde{R}_j^{(t)}}$.

In the arbitration control step, the reliability-weighted action values of the ensemble are computed as follows:
\begin{equation}\label{eq:q_tot}
Q_{\text{total}}(s, a) = \sum_{i=1}^{N} w_i \cdot Q_i(s, a).
\end{equation}

The final action is then selected greedily based on \( Q_{\text{total}} \):
\begin{equation}\label{eq:action}
a^* = \arg\max_a Q_{\text{total}}(s, a).
\end{equation}

This method enables the agents to integrate diverse decision strategy while prioritizing currently the most reliable agent in CRL settings.

\subsubsection{Agency-based sampling}
To address the \emph{curse of diversity} in the ensemble RL, we further extend to the use of reliability to guide the sampling process. We refer to this reliability-based sampling method as \emph{agency-based sampling}, as it is inspired by the notion  ``sense of agency''—allocating experiences to the agent that is the most responsible for generating them (Algorithm \ref{alg:2}).

To allocate transitions (i.e., $\tau = \{s, a, r, s'\}$) across ensemble agents, we first sample transitions using PER~\citep{Schaul2015}. Let \( \tau_j \) denote the \(j\)-th sampled  $n$-step transition (i.e., $\{\tau_j\}_j^{j+n-1}$) from the replay buffer with prioritizing more erroneous transitions~\citep{Sutton1988}. Note that for simplicity, the following description and Algorithm \ref{alg:2} will be presented using one-step transitions.

Each transition is also associated with a loss \( L_j \in \mathbb{R}^N \), where \( N\) is the number of ensemble agents. We convert the loss to reliability as in equation~\eqref{eq:Reliability}, obtains $\mathbf{R}^j\in\mathbb{R}^N$. Note that $\mathbf{R}^j\in\mathbb{R}^N$ is different from $R_i$ of equation~\eqref{eq:Reliability} since temperature parameter $T$ are not the same to favor exploration during training. The value of $T$ in action selection (Algorithm \ref{alg:1}) is smaller than $T$ used in agency-based sampling (Algorithm \ref{alg:2}).

Finally for each $\tau_j$, we draw an agent $i \sim \mathrm{Categorical}(\mathbf{R}^j)$ and assign $\tau_j$ to the agent. For each agent, we retain only the top-$k$ transitions by $p_j$ and perform PER sampling to form $B_i$.

This sampling procedure prevents an agent from being stuck after training on samples with high reliabilities, as the samples make smaller losses. As the sampling starts with $\tau_j$ with PER, the candidate samples are precisely those with high losses, ensuring substantial training.  

\begin{algorithm}[t]
\caption{Agency-based Sampling}
\label{alg:2}
\begin{algorithmic}[1]
    \Statex \textbf{Input:} Replay memory $\mathcal{D}$, temperature $T$, number of agents $N$, batch size $k$, pre-sample size $M$
    \State Initialize temporary buffers $\{\mathcal{D}_{i}^{\mathcal{T}}\}_{i=1}^{N}$

    \Statex \texttt{//Agency Calculation and Assignment}
    \State $\{\tau_j\}_{j=1}^{M}\sim\mathcal{D}$ using PER with priorities
    \For{each transition $\tau_j = (s_j, a_j, r_j, s'_j)$}
        \State Retrieve $\mathbf{L}_j \in \mathbb{R}^N$ for transition $\tau_j$
        \State Compute reliabilities $\mathbf{R}_{j} = \text{softmax}(-\mathbf{L}_{j} / T)$
        \State Sample an agent $i \sim \mathrm{Categorical}(\mathbf{R}_{j})$
        \State Append transition $\tau_j$ with $\mathbf{R}_{j}(i)$ to the buffer $\mathcal{D}_{i}^{\mathcal{T}}$
    \EndFor

    \Statex \texttt{//Agency-based Sampling}
    \For{$i = 1$ to $N$}
        \State Select top-$k$ data $B_i$ from $\mathcal{D}_i^{\mathcal{T}}$ based on reliabilities $\{R_{j,i}\}$
    \EndFor
    \Statex \textbf{Output:} $\{B_i\}_{i=1}^{N}$ 
\end{algorithmic}
\label{alg:2}
\end{algorithm}
\section{Experimental results}
We designed our experiments to answer the following questions:

\begin{enumerate}[label=(\arabic*)]
\item Do optimal performing DQN variants differ in various CRL environments to dynamically select the most reliable agent in the current environmental context? (\ref{optimalDQNVariants})
\item  Is our method of diversified DQN variants and arbitration control in action selection beneficial in CRL environments? (\ref{CRLenv})
\item How crucial is the proposed arbitration control for decision making in discrete control environments? (\ref{AC})
\item Do Agency-based sampling properly sample transitions to attribute their ``sense of agency"? (\ref{AS})
\end{enumerate}

\begin{figure*}[t]
\includegraphics[width=1\linewidth]{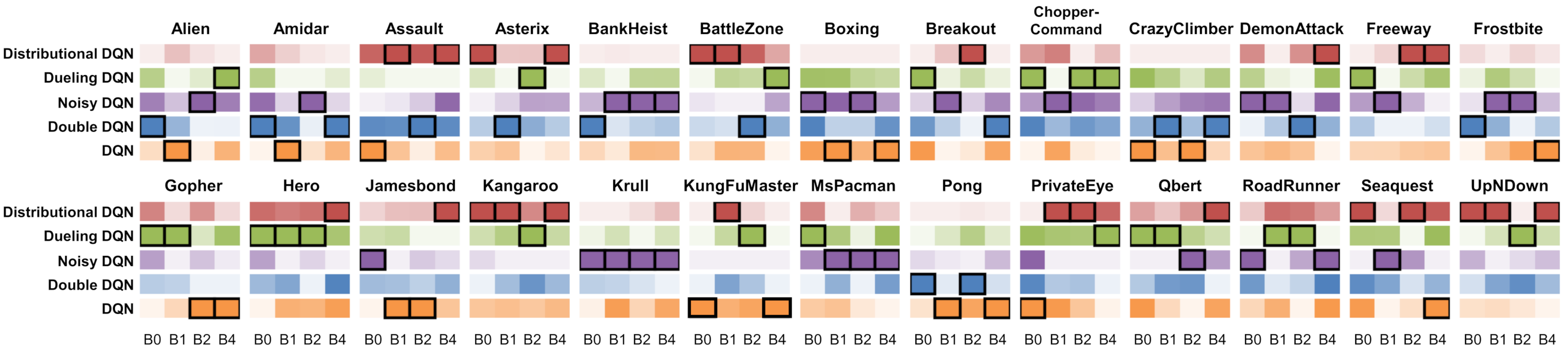}
\vspace{-0.5cm}
\caption{{\bf Performance of DQN variants in each block} Each of DQN variants --DQN, Double DQN, Noisy DQN, Dueling DQN, Distributional DQN-- individually trained on each block. We show mean scores of 30 runs in each block condition normalized to the optimal DQN variant, shown opaque, while other become more translucent as performance decreases.}
\label{fig:sub}
\end{figure*}

\subsection{DQN variants in isolated block conditions}
\label{optimalDQNVariants}

To verify whether different CRL environments favor different optimal agents, we evaluated each variant of DQNs on 4 different condition blocks in 26 Atari tasks~(Figure \ref{fig:sub}). We excluded the condition block 3, as it is the same as the condition block 1. 

We found that the optimal DQN variant differs not only across tasks but also across blocks. No single agent consistently achieved the best performance across all CRL environments. This implies that, when the environment changes, the agent that performs best also changes depending on the context of the task. One can also observe that the change of the optimal agent due to agency becomes more frequent in hard exploration environments such as \textit{Alien}, \textit{Frostbite}, and \textit{Freeway}~\citep{Thangarajah2018}. The exact performance are provided in the Supplementary.

We also provide a theoretical analysis for these empirical findings, focusing on the tabular Q-learning algorithms that underpin these DQN variants to build foundational intuition. The divergence between agents stems from two primary mechanisms: on one hand, agents with inherently biased Bellman operators ($T^{\triangle} \neq T$), such as Double Q-learning, are designed to converge to different stationary points than standard Q-learning. On the other hand, even for theoretically unbiased agents ($T^{\triangle} = T$), the continual learning setting renders the optimal value function ($Q^*$) non-stationary, making performance dependent on the ability to track this moving target. This tracking ability is affected by the variance of the update rule, causing agents with different noise characteristics to exhibit distinct learning dynamics. A detailed formal analysis of both mechanisms is provided in the Supplementary.

\subsection{Performance comparison in CRL}
\label{CRLenv}
\begin{table}[b]
\label{tb:winrate}
\caption{Win rate of ACED-DQN on 26 Atari environments.}
\centering
\begin{tabular}{l|c@{\hspace{8mm}}} 
\hline
Models       & Win Rate \% \\ \hline
ACED-DQN     & 50.5 \%\\
ACED-Rainbow & 22.6 \%\\
SUNRISE      & 16.3 \%\\
Rainbow      & 10.6 \%\\ \hline
\end{tabular}
\end{table}

\begin{figure*}[t]
\centering
\includegraphics[width=1\linewidth]{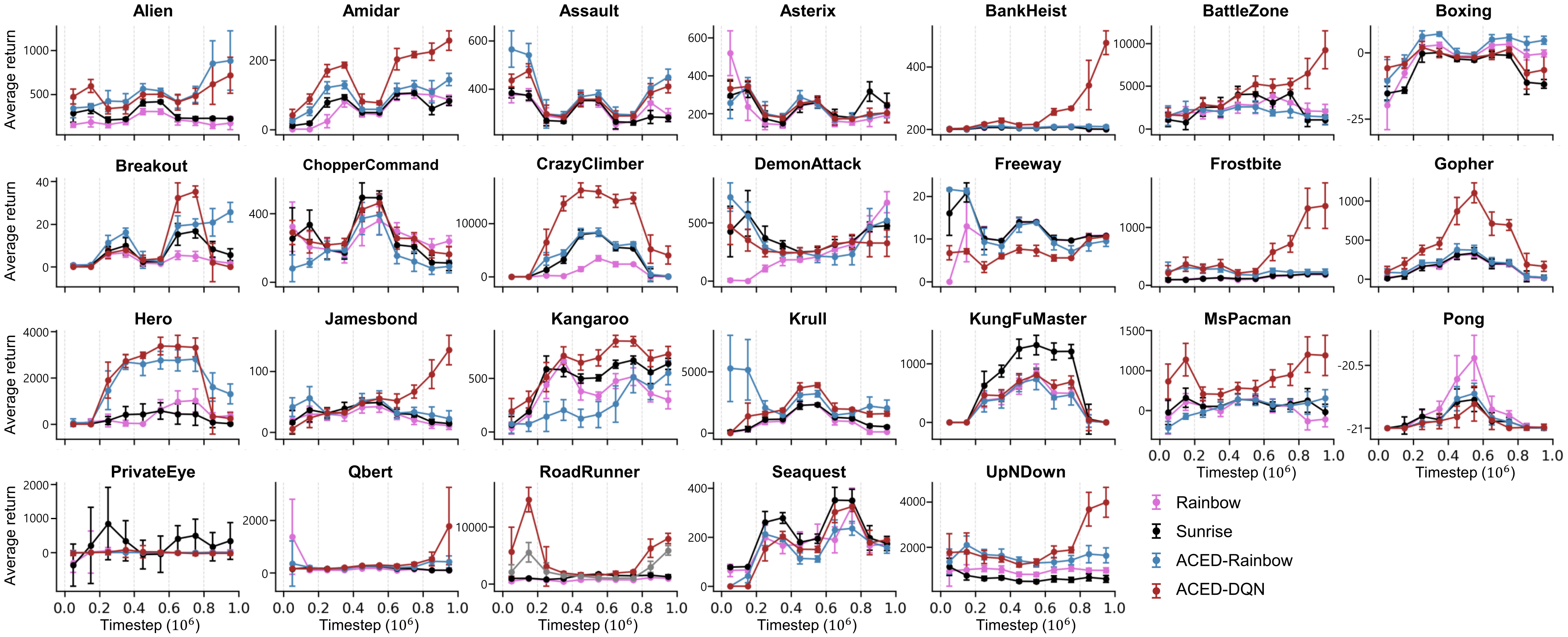}
\caption{{\bf Learning curves of RL agents.} 
Each panel represents averaged episodic return of ACED-DQN, SUNRISE, Rainbow, and ACED-Rainbow. Although there are five environmental context blocks, we plotted two data points for each blocks, resulting 10 markers on each line. Markers indicate mean, and vertical error bars represent standard deviation.}
\label{fig:main}
\end{figure*}

We evaluated the performance of the proposed method (ACED-DQN) in CRL settings against three baselines: Rainbow~\citep{Hessel2017}, an agent that integrates several improvements into the DQN framework; SUNRISE~\citep{Lee2021}, a uniform ensemble method that employs UCB-based exploration with multiple Q-networks; and ACED-Rainbow, an ablation variant of ACED-DQN where Rainbow replaces DQN variants but retains the arbitration control mechanism. Details of hyper-parameter configurations and the implementation of arbitration and sampling mechanisms are provided in the Supplementary.

We found ACED-DQN consistently outperforms Rainbow, SUNRISE, and ACED-Rainbow across a majority of environments, achieving superior performance in 15 out of 26 tasks (Table~\ref{tb:winrate}). In such games, ACED-DQN generally maintains better performance throughout the learning curves (Figure \ref{fig:main}). Interestingly, ACED-DQN also shows strong performance in an initial state of environments (block 0) with dense reward signals, such as Alien, Frostbite, and Kangaroo~\citep{Thangarajah2018}, despite the absence of changes in environmental contexts.

\begin{figure*}[t]
\centering
\includegraphics[width=1\linewidth]{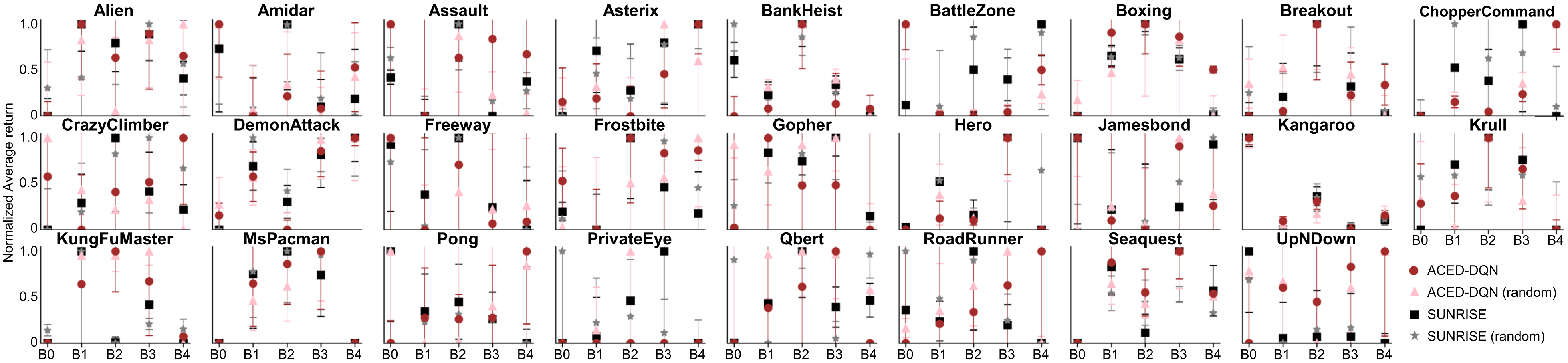}
\caption{\textbf{Performance comparison of RL agents on 26 continual Atari tasks.} 
Each panel corresponds to an Atari environment. Markers denote distinct RL agents, and their performance is evaluated over five consecutive 200k-step blocks (B0-B4). Vertical error bars indicate the standard deviation of the returns within each block.}
\label{fig:ac}
\end{figure*}

\begin{figure*}[t]
\centering
\includegraphics[width=1\linewidth]{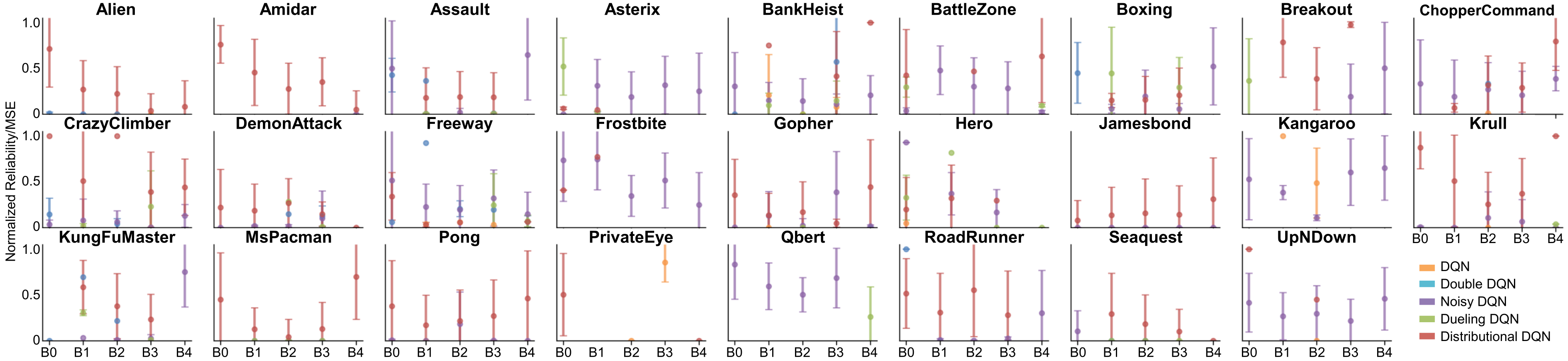}
\caption{\textbf{The effects of Agency-based sampling.} 
Normalized ratio of reliability to MSE in each block is measured to illustrate how ACED-DQN selects different RL agents as an optimal choice. Evaluation is performed across five consecutive 200k-step blocks (B0-B4). Markers and the vertical error bars indicate the mean and standard deviation, respectively. 
}
\label{fig:as}
\end{figure*}

\subsection{Effects of Arbitration control}
\label{AC}
To verify the effectiveness of the proposed arbitration control (Algorithm~\ref{alg:1}), which weights Q-values based on reliability, we compare our method against three baselines: (1) ACED-DQN with random action selection (ACED-DQN (random)); (2) SUNRISE, which selects actions based on the Q-values of a randomly chosen agent (SUNRISE (random)); and (3) standard SUNRISE.

This setup serves as an ablation study for arbitration control, where we compare our method—featuring principled arbitration—with variants that use the same underlying DQN models but apply arbitration randomly, as well as with SUNRISE models. As previously described, SUNRISE follows a fundamentally different approach by training agents, which discourages diversified learning across agents as opposed to our framework. We measured performance in 26 Atari games' block conditions, averaged over 30 evaluation runs.

We found that ACED-DQN generally outperforms the other methods (Figure~\ref{fig:ac}). In terms of win rate, ACED-DQN achieved the highest proportion of wins, accounting for 47.6\% across all games and blocks. Interestingly, ACED-DQN with random agent selection (ACED-DQN (random)) performed similarly, suggesting that the performance gains arise not solely from arbitration control, but also from diversified DQNs during training—likely due to the agency-based sampling that aligns each agent's learning with its own behavior.

Importantly, this result contrasts with prior findings, \emph{curse of diversity}, by \citeauthor{Lin2024}, which highlight that shared replay buffers in ensemble RL can cause interference of value networks and degrade performance due to highly off-policy updates across agents. 

\subsection{Effects of Agency-based sampling}
\label{AS}

\begin{figure}[t]
\centering
\includegraphics[width=1\linewidth]{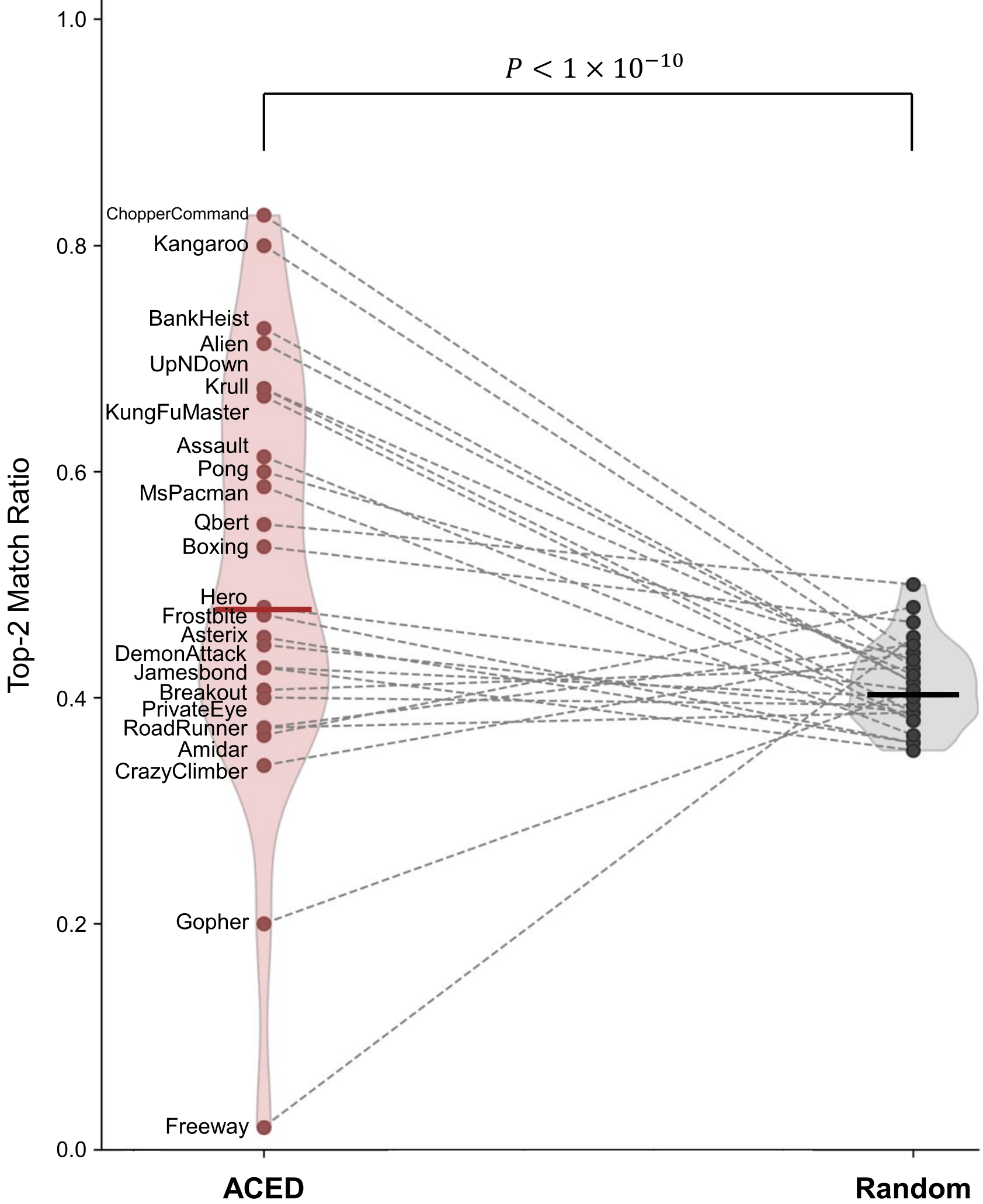}
\caption{\textbf{Arbitration control is well-matched  to the optimal agent.} The optimal agent collected from running individual DQN agents in each block (Figure \ref{fig:main}) is compared with the resulting optimal arbitration controlled agent defined by the normalized reliability over MSE (Figure \ref{fig:as}). The proportion of top-2 matches is reported for each Atari tasks, where each data point is averaged across 5 blocks and 30 evaluation episodes. 
For the Random baseline, ranks are shuffled to demonstrate baseline. A Wilcoxon-signed rank test conducted. 
}
\label{fig:ac_match}
\end{figure}

\subsubsection{Agency-based sampling with arbitration control}
To verify the advantage of agency-based sampling in (Algorithm \ref{alg:2}), we evaluate our sampled transitions during inference for each block in continual Atari games (Figure \ref{fig:as}). For transitions, we calculated a metric, which is a ratio of reliability to MSE. It validates if the most optimal agent (i.e., least MSE), and therefore it can be used to measure degree of optimality of agents. Note that the models are already trained enough, so that an agent's least MSE guarantees that the agent is the best in an ensemble.

We found that Noisy DQN frequently appeared as the most optimal agent across many blocks. However, the optimality metric also fluctuated--not flat--systematically for other agents depending on block condition, indicating that their performance was sensitive to the specific task context.

\subsubsection{Effects of agency-based sampling on CRL environment}
Finally, to empirically verify the effect of agency-based sampling in varying CRL contexts, we perform a Wilcoxon signed-rank test comparing our method with a random sampling. Specifically, we evaluate calculated metrics of ratio of reliability to MSE (Figure \ref{fig:as}) with performance of individual DQN variants in CRL settings (Figure \ref{fig:sub}). From the results of both experiments, we select the DQN variants with the highest return and two agents with the highest reliability from each block, and then measure the Top-2 match ratio for both sampling method (Figure \ref{fig:ac_match}).

The results show that the Top-2 match ratio of the proposed agency-based sampling was 0.51$\pm$ 0.01, while the random permutation sampling achieved 0.41 $\pm$ 0.04. There was significant difference between two methods (Wilcoxon signed-rank test $W=0.0$ and $p= 1.86 \times 10^{-9}$), suggestting agency-based sampling having a stronger correspondence with the actual top-performing agents compared to random permutation sampling.

Thus, we found that the agency-based sampling in ACED-DQN actually weigh an agent that are mostly best performing ones from training DQNs in isolated block conditions. This suggests that arbitraiton control process is not just for emphasizing one that performs the best among a set of agents in an ensemble, but also for training agents with samples they have agency for. 

\section{Discussion}

In this work, we introduced ACED-DQN to address the challenges in CRL. By leveraging an ensemble of DQN variants with diverse value functions and dynamically selecting the most reliable agent through arbitration control, the RL agent effectively adapts to changing environmental contexts. Our experimental results in continual Atari environments highlighted that no single DQN variant consistently performed best across all tasks and conditions, validating the need for dynamic arbitration control. Furthermore, comparative evaluations showed that ACED-DQN outperformed state-of-the-art ensemble (i.e., SUNRISE) as well as its single RL counterpart (i.e., Rainbow DQN), across various CRL settings. These findings emphasize the importance of dynamic arbitration mechanism, enabling ACED-DQN to achieve superior performance compared to state-of-the-art ensemble and single-agent baselines.

A key contribution of our work is addressing the ``curse of diversity," a known challenge in ensemble RL pointed out by \citeauthor{Lin2024}. While ensembles leverage diversity to tackle continual learning, performance can degrade when individual agents are trained with off-policy data from other, behaviorally different members. Our framework mitigates this issue through its arbitration control. By training each agent primarily on samples for which it demonstrates high agency, ACED-DQN ensures that learning remains focused and stable. This targeted approach prevents the performance degradation that can arise from off-policy updates with irrelevant data, effectively resolving the curse of diversity.

Another key strength of our framework is its method of exploration. Rather than relying on explicit stochastic bonuses (e.g., UCB in SUNRISE), ACED-DQN achieves adaptive exploration as a natural consequence of its arbitration mechanism. As reliability shifts among agents --each converging to a distinct stationary point with a unique action value, as shown in our Supplementary -- the system inherently explores different action-value perspectives. This dynamic weighting of diverse policies serves as an implicit exploration driver. However, we acknowledge that a general convergence guarantee for practical DQN remains an open problem, particularly in off-policy CRL settings where the target network is unfixed. Although the variants converge within a bounded error, this is a notable limitation.

Although our evaluation focused on discrete control settings, the framework is generalizable to other domains such as continuous control (e.g., MuJoCo) and multi-task RL. This would involve adapting the reliability metric and arbitration mechanism to corresponding policy and its loss. We specifically chose an ensemble of DQN variants as they represent a well-established and diverse family of value-based algorithms, allowing us to isolate and validate the effectiveness of our arbitration mechanism itself. Likewise, we selected Atari environments as our primary benchmark due to their diverse family of environments and extensive adoption in CRL research.

In conclusion, our work underscores the value of integrating diversified valuation and arbitration processes to enhance adaptability in CRL, inspired by how the human brain addresses similar challenges. By providing a novel perspective on RL system design, this research paves the way for future exploration of arbitration-based ensemble methods to tackle the increasing complexities of continual learning.


\begin{ack}
This work was supported by the National Research Foundation of Korea (NRF) grant funded by the Korea government (MSIT) (RS-2024-00348149), as well as by the MSIT (Ministry of Science and ICT), Korea, under the ICAN (ICT Challenge and Advanced Network of HRD) support program (IITP-2023-RS-2023-00259867) and the ITRC (Information Technology Research Center) support program (IITP-2024-RS-2024-00437102), both supervised by the IITP (Institute for Information \& Communications Technology Planning \& Evaluation). 
\end{ack}


\bibliography{mybibfile}

\end{document}



\begin{frontmatter}


\paperid{2325} 


\title{Supplementary Information}
\end{frontmatter}
\section{Code availability}

The source code for this study is available at \url{https://github.com/iljf/aced_dqn}.

\section{A Theoretical Analysis of Divergence in DQN Variants}

Here, we provide a theoretical analysis of why different Deep Q-Network (DQN) variants, which are empirically observed to produce distinct outcomes, diverge in practice. Our argument unfolds in two main parts.

First, we demonstrate that when a learning algorithm employs a Bellman operator structurally different from the standard one, it provably converges to a different stationary point. This is the case for algorithms such as Double DQN. Although this conclusion may seem self-evident, establishing it formally is a necessary first step in our framework and provides a clear bound on the resulting bias.

Second, and more subtly, we will argue that even when algorithms share the exact same expected Bellman operator, their convergence \emph{dynamics} can differ significantly. Specifically, differences in the variance of the stochastic updates affect the algorithm's trajectory in the value space. In a stationary environment, this leads to different convergence paths; in a more realistic non-stationary environment, this translates in different abilities to track a moving optimal value function. Such difference in tracking capability can cause two theoretically unbiased algorithms to persistently occupy different regions of the value space, effectively leading to practical divergence.

Finally, it is important to note a caveat: the formal analysis presented here is derived in a finite, tabular setting for analytical tractability. It does not constitute a direct proof of convergence for non-linear function approximators like deep neural networks. Nevertheless, this analysis provides a rigorous conceptual framework and clear intuition for understanding the fundamental mechanisms—bias and variance—that underlie the observed performance differences among various DQN agents.

\subsection{Preliminaries}

Let $\mathcal M=(\mathcal S,\mathcal A,P,r,\gamma)$ be a finite Markov Decision Process (MDP) with a state space $\mathcal S$, an action space $\mathcal A$, a transition function $P(s'|s,a)$, a reward function $r(s,a)$, and a discount factor $0<\gamma<1$. To facilitate vector-based analysis, we assume the state-action pairs in $\mathcal S\times\mathcal A$ are ordered, allowing us to treat any Q-function as a vector $Q \in \mathbb R^{|\mathcal S||\mathcal A|}$. The notation $\|\cdot\|_\infty$ denotes the max norm.

\paragraph{Standard Bellman operator.}
The optimal action-value function, $Q^{\!*}$, is the unique solution to the Bellman optimality equation, which is expressed using the Bellman operator, $T: \mathbb R^{|\mathcal S||\mathcal A|} \to \mathbb R^{|\mathcal S||\mathcal A|}$:
\begin{equation}
(TQ)(s,a)=r(s,a)+\gamma\!\sum_{s'}P(s'|s,a)\max_{a'}Q(s',a').
\label{eq:bellman}
\end{equation}
It is well-established that $T$ is a $\gamma$-contraction mapping. By the Banach fixed-point theorem, this guarantees that $T$ has a unique fixed point, $Q^{\!*}$, and that iterating the operator will converge to it.

\subsection{Convergence Analysis}
We analyze the convergence of Q-learning algorithms under the framework of stochastic approximation.

\paragraph{Standard and Variant Updates.}
For a transition $(S_t,A_t,R_t,S_{t+1})$, the standard Q-learning update uses the target $\tau_t^{\textsc{std}} = R_t+\gamma\max_{a'}Q_t(S_{t+1},a')$. A variant algorithm may use a different target, denoted $\tau_t^{\triangle}$. We define the variant's expected Bellman operator, $T^{\triangle}$, as the expectation of its stochastic target, $T^{\triangle}Q=\mathbb E[\tau_t^{\triangle}\mid Q_t=Q]$. The generalized update rule is:
\begin{equation}
Q_{t+1}=Q_t+\alpha_t\bigl[\tau_t^{\triangle} -Q_t(S_t,A_t)\bigr]\mathbf e_{(S_t,A_t)}.
\label{eq:update_var}
\end{equation}

\begin{theorem}\label{thm:biased_unbiased}
Assume (i) every $(s,a)$ is visited infinitely often and (ii) step-sizes satisfy $\sum_t\alpha_t=\infty,\;\sum_t\alpha_t^{2}<\infty$. Then,
\begin{enumerate}[label=(\alph*)]
\item If $T^{\triangle}=T$ (\emph{unbiased}), the recursion converges a.s.\ to $Q^{\!*}$.
\item If $T^{\triangle}\neq T$ (\emph{biased}), the recursion converges a.s.\ to the unique fixed point $Q^{\triangle}$ of $T^{\triangle}$, and $\|Q^{\triangle}-Q^{\!*}\|_\infty \;\le\;\frac{1}{1-\gamma}\sup_{Q}\|TQ-T^{\triangle}Q\|_\infty$.
\end{enumerate}
\end{theorem}
\begin{proof}
Let the noise term be $M_{t+1}=\tau_t^{\triangle}-T^{\triangle}Q_t$. By construction, $M_{t+1}$ is a martingale‑difference sequence. The update becomes $Q_{t+1}=Q_t+\alpha_t\bigl[T^{\triangle}Q_t-Q_t+M_{t+1}\bigr]$, which is a Robbins–Monro recursion. By stochastic‑approximation theory (e.g., \citeauthor{borkar2000ode}), $Q_t$ converges a.s.\ to the solution of $T^{\triangle}Q-Q=0$. This is $Q^{\!*}$ if $T^{\triangle}=T$, and $Q^{\triangle}$ otherwise. The inequality bound follows from the contraction property of the operators.
\end{proof}

\subsection{Biased and Unbiased RL algorithms}

\paragraph{Double Q-learning has a Biased Operator ($T^{\triangle} \neq T$).}
In standard Q-learning, the same value function is used both for action selection and evaluation ($\max_{a'}Q_t(S_{t+1},a')$). Double Q-learning decouples this process using a separate, periodically updated function approximator, $Q_{\text{target}}$. The online function approximator, $Q_{\text{online}}$, selects the action, but the target function approximator evaluates its value:
$\tau_t^{\text{DQ}} = R_t + \gamma Q_{\text{target}}(S_{t+1}, \operatorname{argmax}_{a'} Q_{\text{online}}(S_{t+1}, a'))$.
This structural change leads to a different mathematical expectation for the target. The resulting expected Bellman operator, $T^{\text{DQ}} = \mathbb E[\tau_t^{\text{DQ}} | Q]$, is therefore not equivalent to the standard operator $T$. By Theorem \ref{thm:biased_unbiased}(b), Double Q-learning converges to its own unique stationary point $Q^{\text{DQ}} \neq Q^{\!*}$.

\paragraph{Dueling Q-learning \& Noisy Q-learning are Unbiased ($T^{\triangle} = T$).}
These methods do not alter the expectation of the target value. Dueling Q-learning only restructures the function approximator (by sharing features), while Noisy Q-learning injects zero-mean noise into parameters. In expectation, their operators are equivalent to the standard Bellman operator, $T$.

\subsection{Divergence emerges from Learning Dynamics and Variance}

Even when algorithms are unbiased ($T^{\triangle} = T$), they can diverge due to different learning dynamics, especially in continual reinforcement learning (i.e., non-stationary) environments. This divergence is driven by the variance of the update noise term, $M_{t+1}$.

In a non-stationary task, the optimal value function $Q^{\!*}_t$ changes over time. The algorithm's goal is to track this moving target. The tracking error, $\delta_t=\|Q_t-Q^{\!*}_t\|_\infty$, evolves according to:
\begin{equation}
\delta_{t+1}\le\Bigl(1-\tfrac{\alpha_t}{1-\gamma}\Bigr)\delta_t +\alpha_t\bigl(\beta_t+\|M_{t+1}\|\bigr),
\label{eq:track_short}
\end{equation}
where $\beta_t=\|Q^{\!*}_{t+1}-Q^{\!*}_t\|_\infty$ is the environmental drift. This inequality reveals that the tracking error is driven by the magnitude of the update noise, $\|M_{t+1}\|$.

\paragraph{Mathematical Justification.}
We can formalize the claim that larger Variance of $M_{t+1}$ leads to a larger tracking error by analyzing the expectation of Eq. \eqref{eq:track_short}. Taking the expectation, we find that the expected steady-state error, $\delta_{ss} = \lim_{t\to\infty} \mathbb E[\delta_t]$, is bounded by:
\[
\delta_{ss} \le (1-\gamma)\left(\mathbb E[\beta_t] + \mathbb E\left[\|M_{t+1}\|\right]\right)
\]
This shows that the persistent tracking error is directly influenced by the expected magnitude of the noise, $\mathbb E\left[\|M_{t+1}\|\right]$. We can connect this magnitude to the variance using Jensen's inequality. For the L2-norm, it can be shown that $\left(\mathbb E\left[\|M_{t+1}\|_2\right]\right)^2 \le \mathbb E\left[\|M_{t+1}\|_2^2\right] = \operatorname{Var}(M_{t+1})$. This implies that a larger noise variance leads to a larger expected noise magnitude.

In practice, algorithms that inherently generate high-variance updates will suffer from larger persistent tracking errors in dynamic environments.
\begin{itemize}
    \item \textbf{Noisy Q-learning} introduces parameter noise, which is an additional source of randomness in $\tau_t$, thus inflating $\operatorname{Var}(M_{t+1})$.
    \item \textbf{Prioritized Replay} prioritizes high-error transitions, which typically exhibit high-variance samples for the update, again inflating $\operatorname{Var}(M_{t+1})$.
\end{itemize}
Therefore, even though these methods are theoretically unbiased, their different noise characteristics result in differing adaptive capabilities. In a non-stationary setting, this difference in tracking ability leads them to maintain divergent value functions, explaining their divergence in practice.

\section{Changes in Environmental contexts in continual Atari games}

\subsection{Changes in objectives}
We show the objectives of 26 Atari tasks perturbed by changing the reward functions (Table).

\begin{table*}[h]
\label{tb:rm}
\caption{Perturbed objectives for Atari tasks}
\centering
\begin{tabular}{l|c|c@{\hspace{8mm}}} 
\hline
Environments&  Default objective: 0&Modified objective: 1\\ \hline
Alien&  Destroy eggs avoiding aliens&Track down aliens\\
Amidar&  Color the entire maze&Walk endlessly\\
Assault&  Destroy all enemies that approaches&Focus destroying smaller drones\\
 Asterix& Collect objects&Skip object once in a while\\
 BankHeist& Focus robbing a bank before the gas rounds out&Destroy police cars dropping dynamite\\
 BattleZone& Destroy enemy vehicles&Focus destroying airships\\
 Boxing& Knock out your opponent&Punch and jabs are perturbed\\
 ChopperCommand& Knockout enemy aircraft and defend truck convoy&Knockout helicopters while avoiding jets\\
 CrazyClimber& Climb to the top of the building as fast possible&Try to get hit by every obsticle\\
 DemonAttack& Destroy enimies&Shoot at foot demons only\\
 Freeway& Cross all the lane&Cross half the lane\\
 Frostbite& Jump on ice block untill the igloo is built&Jump endlessly untill the temperature drops out\\
 Gopher& Protect crops from gopher&Try to eliminate gopher\\
 Hero& Rescue miners&Focus using all the dynamite\\
 Jamesbond& Shoot at the enemies to survive&Try to dodge enemy contact\\
 Kangaroo& Rescue baby kangaroo&Punching monkeys\\
 Krull& Rescue princess and destroy the beast&Try to gain points before reaching cacoon\\
 KungFuMaster& Rescue princess by defeating enemies&Scores for Kick and Punch perturbed\\
 MsPacman& Collect all the pellets&Chase down ghosts\\
 Pong& Make points by deflecting the ball into the goal&Stochastic points\\
 PrivateEye& Find evidence and stolen goods&Getting hit by obsticles will grant more points\\
 Qbert& Chage the color of all the cubes&avoid nasty creatures\\
 RoadRunner& Focus collecting seeds&Focus killing the coyote\\
 Seaquest& Retrieve divers&Kill fish\\
UpNDown&  Collect prizes and eliminate opponents&Losse a point each time you move\\ \hline
\end{tabular}
\end{table*}

\subsection{Changes in probability of random actions}
We implement action randomness to create stochastic action selection. To inject stochasticity during action selection, each action \( a \in \mathcal{A} \) is decomposed into two components: a directional component \( d \in \mathcal{D} \), and a firing flag \( f \in \{0, 1\} \). During evaluation, each component is independently perturbed with probability \( \epsilon \), and the final executed action \( \tilde{a} \) is constructed as:
\[
\begin{aligned}\label{eq:action}
d' &=
\begin{cases}
\text{sample from } \mathcal{D} & \text{with probability } \epsilon \\
d & \text{with probability } 1 - \epsilon
\end{cases} \\\\[4pt]
f' &=
\begin{cases}
\text{sample from } \{0,1\} & \text{with probability } \epsilon \\
f & \text{with probability } 1 - \epsilon
\end{cases} \\\\[4pt]
\tilde{a} &= \text{encode}(d', f')
\end{aligned}
\]
where, \( \text{encode}() \) maps the perturbed direction and fire flag back to discrete action index in the environment's action space. (i.e. \( \tilde{a} = d' + 8 \cdot f' + 2 \)  for an 18-dimensional action space)

\section{Implementation details and hyper-parameters}

In this section, we provide implementation details of ensemble algorithms and hyper-parameter used in this work.

\subsection{DQN variants}
\paragraph{DQN}
We follow the standard implementation of DQN as proposed by \citeauthor{Mnih2013}\citep{Mnih2013}. The network has three convolutional layers: the first layer uses 32 filters of size $8 \times 8$ with stride 4, the second layer uses 64 filters of size $4 \times 4$ with stride 2, and the third layer uses 64 filters of size $3 \times 3$ with stride 1.  In addition, the network has the hidden layer with 512 units and number of output layers equal to the number of actions in the task.

\paragraph{Double DQN}
Double DQN shares the same network architecture as DQN. The only difference is the target Q-value computation, where the action is selected using the online network and evaluated using the target network  as:
\begin{equation}\label{eq:ddqn}
y_t = r_t + \gamma Q_{\theta^-}\left(s_{t+1}, \arg\max_{a'} Q_\theta(s_{t+1}, a')\right).
\end{equation}
where $\theta$ and $\theta^-$ denote the parameters of the online and target networks.

\paragraph{Dueling DQN}
Instead of directly estimating Q-values, the network is decomposed into two separate output streams: one estimating the state-value function \(V(s)\), and the other estimating the advantage function \(A(s, a)\). These two components are combined to produce the final Q-value as follows:
\begin{equation}\label{eq:dueling}
Q(s, a) = V(s) + \left(A(s, a) - \frac{1}{|\mathcal{A}|} \sum_{a'} A(s, a')\right).
\end{equation}
This modification is made after the final convolutional layer.

\paragraph{Noisy DQN}
Noisy DQN  extends the standard DQN architecture by replacing the deterministic fully connected layers in the network with stochastic, learnable Noisy Linear layers~\citep{Fortunato2017}. All other components remain consistent with the original DQN implementation.

For our method of Noisy DQN, exploration is achieved through factorized Gaussian noise in the weights and biases of the linear layers. Each Noisy Linear layer is defined as:
\begin{equation}\label{eq:noisy}
y = (\mu_W + \sigma_W \odot \epsilon_W) x + (\mu_b + \sigma_b \odot \epsilon_b).
\end{equation}
where \(\mu_W, \sigma_W\) and \(\mu_b, \sigma_b\) are learnable parameters, and \(\epsilon_W, \epsilon_b\) are noise variables. When using Noisy DQN, we acted fully greedily with a value of 0.5 for the hyper-parameter used to initialize the weights in the noisy stream.

\paragraph{Distributional DQN}
Distributional DQN modifies the standard DQN architecture by modeling the full distribution over future returns instead of estimating only their expected value~\citep{Bellemare2017}. This change affects both the network output and the loss function.

In Distributional DQN, the Q-value for a given state-action pair is approximated by a categorical distribution over a fixed set of discrete support values (atoms), spaced linearly between a minimum and maximum return range \([v_{\min}, v_{\max}]\). Instead of outputting a scalar Q-value, the network predicts a probability mass function over these atoms for each action. The Q-value can be recovered as the expected value of the distribution:
\begin{equation}\label{eq:distri}
Q(s, a) = \sum_{i=1}^N z_i \cdot p_i(s, a).
\end{equation}
where \(z_i\) are the support values and \(p_i\) are the predicted probabilities for each atom. The loss is computed using a cross-entropy divergence between the projected target distribution and the predicted distribution.

Since other DQN variants estimate scalar Q-values and compute prediction error using mean squared error (MSE),  to enable consistent arbitration control across ensemble members based on reliability, we convert the prediction from Distributional DQN into a scalar Q-value by taking the expectation over the support: 
\begin{equation}\label{eq:CE}
q = \sum_{z \in \mathcal{Z}} z \cdot p(z \mid s, a),
\end{equation}
where \(\mathcal{Z} \) denotes the discrete support of returns and \( p(z \mid s, a) \) is the predicted probability for support value \( z \) at state-action pair \( (s, a) \). The target Q-value \( q_{\text{target}} \) is computed similarly from the target distributional network. We then apply an MSE loss between \( q \) and \( q_{\text{target}} \) to align with the loss formulation of other ensemble agents:
\begin{equation}\label{eq:scalar}
\mathcal{L}_{\text{mse}} = \| q - q_{\text{target}} \|_2^2,
\end{equation}
Since the magnitude of this loss can differ from the MSE losses of scalar Q-value models, we apply a rescaling factor \( \alpha \) to maintain comparable loss ranges:
\begin{equation}\label{eq:normaldistri}
\mathcal{L}_{\text{normalized}} = \alpha \cdot \mathcal{L}_{\text{mse}}.
\end{equation}
This normalization ensures that the reliability-based arbitration is not biased due to inherently different loss scales across ensemble members.

\subsection{Hyper-parameters}
We initialize hyper-parameters that of the same values introduced by DQN and its variants for preprocessing of environment frames (Table \ref{tb:prepara}) and additional hyper-parameters that ACED-DQN inherits from DQN (Table \ref{tb:add-para}).
\begin{table}[b]
\caption{Preprocessing of environment frames}
\label{tb:prepara}
\centering
\begin{tabular}{l|c@{\hspace{8mm}}} 
\hline
Hyper-parameter& value\\ \hline
Grey-scaling& True\\
Observation down-sampling& (84, 84)\\
Frames stacked& 4\\
 Action repetitions&4\\
 Reward clipping&[-1,1]\\
 Terminal on loss of life&True\\
Max frames per episode& 108K\\ \hline
\end{tabular}
\end{table}

\begin{table}[t]
\caption{Additional hyper-parameters}
\label{tb:add-para}
\centering
\begin{tabular}{l|c@{\hspace{8mm}}} 
\hline
Hyper-parameter& value\\ \hline
Q-network : channels& 32, 64, 64\\
Q-network : filter size& 8$\times$8, 4$\times$4, 3$\times3$\\
Q-network : stride& 4, 2, 1\\
 Q-network : hidden units&512\\
 Q-network : output units&Number of actions\\
 Discount factor&0.99\\
Minibatch size& 32\\ \hline
\end{tabular}
\end{table}
The hyper-parameters for ACED-DQN are identical across all 26 Atari tasks (Table \ref{tb:acedpara}).

\begin{table}[t]
\caption{ACED-DQN hyper-parameters}
\label{tb:acedpara}
\centering
\begin{tabular}{l|c@{\hspace{8mm}}} 
\hline
Hyper-parameter& value\\ \hline
Min history to start learning& 80K frames\\
Target Network Period& 32K frames\\
 Cycle cosine learning rate&$0.001\rightarrow 0.0$ \, \text{for} \, 200K\\
Adam $\epsilon$& $1.5\times10^{-4}$\\
 Prioritization type &proportional\\
 Prioritization exponent&0.5\\
 Prioritization importance sampling&0.4\\
 Multi-step returns&3\\
 Distributional atoms&51\\
 Momentum update $\gamma$&0.6\\
 Distributional min/max values&[-10,10]\\
 Arbitration control temperature $T$&0.3\\
Agency sampling temperature $T$& 0.8\\
Reliability clipping $R_{min},R_{max}$& [0.2,0.5]\\ \hline

\end{tabular}
\end{table}

\subsection{Experimental Details}
\paragraph{DQN variants in isolated block conditions}
We evaluate the performance of the DQN variants under isolated block conditions across 26 Atari tasks, based on modifications of the environments. For training, we consider DQN, Double DQN, Noisy DQN, Dueling DQN, and Distributional DQN, without any modification to the hyperparameters or architectures as specified in the original work. Each agent is trained for three runs per block and reset after completing training in each block. The optimal parameters are selected to achieve the best performance in the current environment. For evaluation, performance is measured over 30 independent evaluation episodes for each block condition and each DQN variant.

\paragraph{Atari with CRL environments}
For Rainbow and SUNRISE, we use the original hyperparameter configurations as reported in the respective implementations. For our method, the arbitration control mechanism employs a temperature of $T=0.3$ and a momentum parameter of $\gamma=0.6$ to dynamically adjust reliability scores. In agency-based sampling, we set the temperature to $T=0.8$ to regulate reliability (Table~\ref{tb:acedpara}). This configuration allows agents to select actions based on the most reliable estimate in the current environmental context, while the sampling mechanism prioritizes transitions consistent with each agent’s reliability.

\paragraph{Ablation study}
We conduct an experimental evaluation of the effectiveness of arbitration control and agency-based sampling. During training, our method employs the proposed arbitration control to weigh Q-values according to agent reliability, while SUNRISE uses UCB-based exploration as originally proposed. Each block is trained for 200k environment steps sequentially. The ensembles are then evaluated over 50 episodes, with our method updating reliability on a per-step basis. In contrast, SUNRISE evaluates the ensembles using random agent selection for policy execution. Performance is measured over 30 independent evaluation runs.

\begin{table*}[h]
\caption{\textbf{Performance of DQN variants (1/2)} with modified block conditions on 26 Atari tasks. The results show the optimal performing DQN variant --DQN, Double DQN, Noisy DQN, Dueling DQN, and Distributional DQN--  with  mean and standard deviation averaged thirty runs in each block conditions.}
\label{tb:1_2}
\centering
\begin{tabular}{l|c|@{\hspace{1mm}}llllllllll}
\hline
 Environments&  Blocks&  \multicolumn{2}{c}{DQN}&  \multicolumn{2}{c}{Double DQN}&  \multicolumn{2}{c}{Noisy DQN}&  \multicolumn{2}{c}{Dueling DQN}&  \multicolumn{2}{c}{Distributional DQN}\\ \hline
 Alien&  0
&   \multicolumn{2}{c}{341.7 $\pm$  181.0}&   \multicolumn{2}{c}{722.3 $\pm$ 694.6}&   \multicolumn{2}{c}{617.3 $\pm$ 258.3}&   \multicolumn{2}{c}{555.3 $\pm$ 206.4}&   \multicolumn{2}{c}{218.0 $\pm$ 22.1}\\
 &  1
&   \multicolumn{2}{c}{441.5 $\pm$ 783.8}&   \multicolumn{2}{c}{340.2 $\pm$ 574.9}&   \multicolumn{2}{c}{291.8 $\pm$ 323.1}&   \multicolumn{2}{c}{218.0 $\pm$ 293.8}&   \multicolumn{2}{c}{282.0 $\pm$ 584.1}\\
 &  2
&   \multicolumn{2}{c}{297.3 $\pm$ 121.5}&   \multicolumn{2}{c}{287.0 $\pm$ 122.9}&   \multicolumn{2}{c}{502.0 $\pm$ 212.3}&   \multicolumn{2}{c}{333.0 $\pm$ 149.4}&   \multicolumn{2}{c}{303.0 $\pm$ 156.6}\\
 &  4&   \multicolumn{2}{c}{711.7 $\pm$ 1122.5}&   \multicolumn{2}{c}{121.3 $\pm$ 34.7}&   \multicolumn{2}{c}{807.8 $\pm$ 948.7}&   \multicolumn{2}{c}{976.0 $\pm$ 1255.8}&   \multicolumn{2}{c}{116.0 $\pm$ 23.0}\\ \hline
 Amidar&  0
&   \multicolumn{2}{c}{50.4 $\pm$ 33.7}&   \multicolumn{2}{c}{105.6 $\pm$ 47.8}&   \multicolumn{2}{c}{101.6 $\pm$ 25.0}&   \multicolumn{2}{c}{87.4 $\pm$ 22.6}&   \multicolumn{2}{c}{75.3 $\pm$ 2.1}\\
 &  1
&   \multicolumn{2}{c}{131.8 $\pm$ 49.0}&   \multicolumn{2}{c}{127.2 $\pm$ 42.0}&   \multicolumn{2}{c}{106.8 $\pm$ 57.4}&   \multicolumn{2}{c}{100.5 $\pm$ 41.4}&   \multicolumn{2}{c}{106.4 $\pm$ 46.8}\\
 &  2
&   \multicolumn{2}{c}{51.4 $\pm$ 19.2}&   \multicolumn{2}{c}{48.8 $\pm$ 17.6}&   \multicolumn{2}{c}{64.0 $\pm$ 29.5}&   \multicolumn{2}{c}{48.4 $\pm$ 16.0}&   \multicolumn{2}{c}{49.1 $\pm$ 15.7}\\
 &  4&   \multicolumn{2}{c}{153.8 $\pm$ 18.0}&   \multicolumn{2}{c}{171.9 $\pm$ 40.5}&   \multicolumn{2}{c}{127.3 $\pm$ 41.3}&   \multicolumn{2}{c}{127.7 $\pm$ 26.7}&   \multicolumn{2}{c}{117.7 $\pm$ 37.1}\\ \hline
 Assault&  0
&   \multicolumn{2}{c}{626.5 $\pm$ 74.8}&   \multicolumn{2}{c}{613.2 $\pm$ 105.5}&   \multicolumn{2}{c}{481.6 $\pm$ 106.9}&   \multicolumn{2}{c}{504.7 $\pm$ 161.7}&   \multicolumn{2}{c}{607.6 $\pm$ 116.9}\\
 &  1
&   \multicolumn{2}{c}{338.5 $\pm$ 92.9}&   \multicolumn{2}{c}{398.0 $\pm$ 93.5}&   \multicolumn{2}{c}{262.0 $\pm$ 75.8}&   \multicolumn{2}{c}{251.5 $\pm$ 66.1}&   \multicolumn{2}{c}{418.5 $\pm$ 113.1}\\
 &  2
&   \multicolumn{2}{c}{354.2 $\pm$ 87.6}&   \multicolumn{2}{c}{457.4 $\pm$ 140.8}&   \multicolumn{2}{c}{376.6 $\pm$ 80.0}&   \multicolumn{2}{c}{347.9 $\pm$ 92.8}&   \multicolumn{2}{c}{450.4 $\pm$ 132.3}\\
 &  4&   \multicolumn{2}{c}{401.0 $\pm$ 103.8}&   \multicolumn{2}{c}{454.0 $\pm$ 57.8}&   \multicolumn{2}{c}{462.0 $\pm$ 80.1}&   \multicolumn{2}{c}{307.0 $\pm$ 95.7}&   \multicolumn{2}{c}{468.5 $\pm$ 43.9}\\ \hline
 Asterix&  0
&   \multicolumn{2}{c}{660.0 $\pm$ 20.0}&   \multicolumn{2}{c}{635.0 $\pm$ 64.7}&   \multicolumn{2}{c}{468.3 $\pm$ 259.0}&   \multicolumn{2}{c}{620.0 $\pm$ 119.4}&   \multicolumn{2}{c}{965.0 $\pm$ 218.4}\\
 &  1
&   \multicolumn{2}{c}{228.3 $\pm$ 131.7}&   \multicolumn{2}{c}{271.3 $\pm$ 135.0}&   \multicolumn{2}{c}{202.0 $\pm$ 116.9}&   \multicolumn{2}{c}{183.7 $\pm$ 79.9}&   \multicolumn{2}{c}{205.3 $\pm$ 104.7}\\
 &  2
&   \multicolumn{2}{c}{241.7 $\pm$ 91.4}&   \multicolumn{2}{c}{298.3 $\pm$ 136.3}&   \multicolumn{2}{c}{291.7 $\pm$ 142.6}&   \multicolumn{2}{c}{333.3 $\pm$ 147.9}&   \multicolumn{2}{c}{265.0 $\pm$ 214.9}\\
 &  4&   \multicolumn{2}{c}{317.0 $\pm$ 132.5}&   \multicolumn{2}{c}{369.0 $\pm$ 82.8}&   \multicolumn{2}{c}{420.0 $\pm$ 32.9}&   \multicolumn{2}{c}{274.7 $\pm$ 108.7}&   \multicolumn{2}{c}{546.0 $\pm$ 120.8}\\ \hline
 BankHeist&  0
&   \multicolumn{2}{c}{31.0 $\pm$ 5.4}&   \multicolumn{2}{c}{62.3 $\pm$ 30.9}&   \multicolumn{2}{c}{36.0 $\pm$ 14.0}&   \multicolumn{2}{c}{29.0 $\pm$ 8.3}&   \multicolumn{2}{c}{17.3 $\pm$ 10.6}\\
 &  1
&   \multicolumn{2}{c}{38.8 $\pm$ 32.7}&   \multicolumn{2}{c}{38.8 $\pm$ 27.1}&   \multicolumn{2}{c}{68.5 $\pm$ 24.4}&   \multicolumn{2}{c}{34.8 $\pm$ 21.5}&   \multicolumn{2}{c}{35.7 $\pm$ 21.0}\\
 &  2
&   \multicolumn{2}{c}{17.3 $\pm$ 9.6}&   \multicolumn{2}{c}{12.7 $\pm$ 7.7}&   \multicolumn{2}{c}{22.7 $\pm$ 19.7}&   \multicolumn{2}{c}{16.7 $\pm$ 13.7}&   \multicolumn{2}{c}{8.3 $\pm$ 14.9}\\
 &  4&   \multicolumn{2}{c}{59.2 $\pm$ 46.3}&   \multicolumn{2}{c}{48.3 $\pm$ 27.3}&   \multicolumn{2}{c}{78.3 $\pm$ 8.5}&   \multicolumn{2}{c}{58.2 $\pm$ 6.6}&   \multicolumn{2}{c}{29.2 $\pm$ 18.4}\\ \hline
 BattleZone&  0
&   \multicolumn{2}{c}{3300.0 $\pm$ 2267.9}&   \multicolumn{2}{c}{3633.3 $\pm$ 2774.7}&   \multicolumn{2}{c}{2066.7 $\pm$ 1860.7}&   \multicolumn{2}{c}{1233.3 $\pm$ 1782.9}&   \multicolumn{2}{c}{12066.7 $\pm$ 4057.4}\\
 &  1
&   \multicolumn{2}{c}{6233.3 $\pm$ 3792.0}&   \multicolumn{2}{c}{2350.0 $\pm$ 3304.4}&   \multicolumn{2}{c}{1233.3 $\pm$ 2076.6}&   \multicolumn{2}{c}{5466.7 $\pm$ 4795.7}&   \multicolumn{2}{c}{8483.3 $\pm$ 5396.7}\\
 &  2
&   \multicolumn{2}{c}{7900.0 $\pm$ 5509.7}&   \multicolumn{2}{c}{10066.7 $\pm$ 6683.0}&   \multicolumn{2}{c}{3333.3 $\pm$ 2412.9}&   \multicolumn{2}{c}{6800.0 $\pm$ 5594.0}&   \multicolumn{2}{c}{9333.3 $\pm$ 4422.2}\\
 &  4&   \multicolumn{2}{c}{1816.7 $\pm$ 3056.4}&   \multicolumn{2}{c}{4316.7 $\pm$ 4052.7}&   \multicolumn{2}{c}{4116.7 $\pm$ 4308.3}&   \multicolumn{2}{c}{6083.3 $\pm$ 6857.0}&   \multicolumn{2}{c}{3716.7 $\pm$ 2809.8}\\ \hline
 Boxing&  0
&   \multicolumn{2}{c}{-4.0 $\pm$ 3.7}&   \multicolumn{2}{c}{-2.6 $\pm$ 5.3}&   \multicolumn{2}{c}{-2.3 $\pm$ 2.8}&   \multicolumn{2}{c}{-3.8 $\pm$ 4.3}&   \multicolumn{2}{c}{-19.6 $\pm$ 2.7}\\
 &  1
&   \multicolumn{2}{c}{12.6 $\pm$ 9.2}&   \multicolumn{2}{c}{5.4 $\pm$ 9.2}&   \multicolumn{2}{c}{12.5 $\pm$ 11.3}&   \multicolumn{2}{c}{11.6 $\pm$ 8.1}&   \multicolumn{2}{c}{2.5 $\pm$ 11.4}\\
 &  2
&   \multicolumn{2}{c}{-1.5 $\pm$ 6.5}&   \multicolumn{2}{c}{-3.4 $\pm$ 4.8}&   \multicolumn{2}{c}{-0.1 $\pm$ 4.0}&   \multicolumn{2}{c}{-2.0 $\pm$ 7.1}&   \multicolumn{2}{c}{-4.8 $\pm$ 8.6}\\
 &  4&   \multicolumn{2}{c}{16.6 $\pm$ 9.6}&   \multicolumn{2}{c}{12.5 $\pm$ 6.8}&   \multicolumn{2}{c}{5.8 $\pm$ 6.3}&   \multicolumn{2}{c}{3.4 $\pm$ 5.8}&   \multicolumn{2}{c}{-10.5 $\pm$ 10.5}\\ \hline
 Breakout&  0
&   \multicolumn{2}{c}{6.4 $\pm$ 5.1}&   \multicolumn{2}{c}{5.7 $\pm$ 4.6}&   \multicolumn{2}{c}{2.9 $\pm$ 1.1}&   \multicolumn{2}{c}{6.6 $\pm$ 4.9}&   \multicolumn{2}{c}{1.9 $\pm$ 3.6}\\
 &  1
&   \multicolumn{2}{c}{9.4 $\pm$ 16.2}&   \multicolumn{2}{c}{12.2 $\pm$ 16.8}&   \multicolumn{2}{c}{18.1 $\pm$ 16.8}&   \multicolumn{2}{c}{10.4 $\pm$ 17.2}&   \multicolumn{2}{c}{11.7 $\pm$ 17.9}\\
 &  2
&   \multicolumn{2}{c}{1.8 $\pm$ 3.0}&   \multicolumn{2}{c}{1.7 $\pm$ 2.9}&   \multicolumn{2}{c}{2.0 $\pm$ 2.6}&   \multicolumn{2}{c}{2.2 $\pm$ 3.3}&   \multicolumn{2}{c}{2.8 $\pm$ 4.3}\\
 &  4&   \multicolumn{2}{c}{29.2 $\pm$ 17.7}&   \multicolumn{2}{c}{32.2 $\pm$ 16.4}&   \multicolumn{2}{c}{27.5 $\pm$ 5.7}&   \multicolumn{2}{c}{16.8 $\pm$ 19.8}&   \multicolumn{2}{c}{14.9 $\pm$ 11.2}\\ \hline
 ChopperCommand&  0
&   \multicolumn{2}{c}{720.0 $\pm$ 205.6}&   \multicolumn{2}{c}{836.7 $\pm$ 231.6}&   \multicolumn{2}{c}{793.3 $\pm$ 244.9}&   \multicolumn{2}{c}{840.0 $\pm$ 245.8}&   \multicolumn{2}{c}{783.3 $\pm$ 235.3}\\
 &  1
&   \multicolumn{2}{c}{500.0 $\pm$ 170.3}&   \multicolumn{2}{c}{408.3 $\pm$ 155.9}&   \multicolumn{2}{c}{538.7 $\pm$ 385.8}&   \multicolumn{2}{c}{275.0 $\pm$ 164.8}&   \multicolumn{2}{c}{458.0 $\pm$ 175.0}\\
 &  2
&   \multicolumn{2}{c}{716.7 $\pm$ 198.5}&   \multicolumn{2}{c}{820.0 $\pm$ 261.3}&   \multicolumn{2}{c}{843.3 $\pm$ 340.3}&   \multicolumn{2}{c}{850.0 $\pm$ 294.1}&   \multicolumn{2}{c}{703.3 $\pm$ 237.3}\\
 &  4&   \multicolumn{2}{c}{311.7 $\pm$ 173.6}&   \multicolumn{2}{c}{434.0 $\pm$ 91.5}&   \multicolumn{2}{c}{458.3 $\pm$ 69.6}&   \multicolumn{2}{c}{492.7 $\pm$ 130.2}&   \multicolumn{2}{c}{367.0 $\pm$ 88.0}\\ \hline
 CrazyClimber&  0
&   \multicolumn{2}{c}{14950.0 $\pm$ 1681.4}&   \multicolumn{2}{c}{11000.0 $\pm$ 0.0}&   \multicolumn{2}{c}{7950.0 $\pm$ 1972.1}&   \multicolumn{2}{c}{14773.3 $\pm$ 1368.0}&   \multicolumn{2}{c}{5903.3 $\pm$ 2649.5}\\
 &  1
&   \multicolumn{2}{c}{13084.0 $\pm$ 2162.3}&   \multicolumn{2}{c}{16851.3 $\pm$ 5602.3}&   \multicolumn{2}{c}{13126.7 $\pm$ 3073.1}&   \multicolumn{2}{c}{13990.0 $\pm$ 3330.0}&   \multicolumn{2}{c}{8219.0 $\pm$ 2685.3}\\
 &  2
&   \multicolumn{2}{c}{17270.0 $\pm$ 4336.0}&   \multicolumn{2}{c}{14853.3 $\pm$ 3752.0}&   \multicolumn{2}{c}{16276.7 $\pm$ 4901.9}&   \multicolumn{2}{c}{13973.3 $\pm$ 3950.5}&   \multicolumn{2}{c}{12720.0 $\pm$ 3162.4}\\
 &  4&   \multicolumn{2}{c}{5040.0 $\pm$ 277.7}&   \multicolumn{2}{c}{9147.7 $\pm$ 3183.2}&   \multicolumn{2}{c}{8241.3 $\pm$ 1295.3}&   \multicolumn{2}{c}{7522.7 $\pm$ 452.1}&   \multicolumn{2}{c}{3243.3 $\pm$ 1595.4}\\ \hline
 DemonAttack&  0
&   \multicolumn{2}{c}{635.3 $\pm$ 379.0}&   \multicolumn{2}{c}{229.0 $\pm$ 163.2}&   \multicolumn{2}{c}{1025.7 $\pm$ 435.9}&   \multicolumn{2}{c}{739.8 $\pm$ 346.7}&   \multicolumn{2}{c}{723.2 $\pm$ 558.3}\\
 &  1
&   \multicolumn{2}{c}{408.2 $\pm$ 342.3}&   \multicolumn{2}{c}{361.5 $\pm$ 210.1}&   \multicolumn{2}{c}{472.2 $\pm$ 255.6}&   \multicolumn{2}{c}{314.5 $\pm$ 173.2}&   \multicolumn{2}{c}{297.0 $\pm$ 214.3}\\
 &  2
&   \multicolumn{2}{c}{211.5 $\pm$ 173.8}&   \multicolumn{2}{c}{254.5 $\pm$ 152.5}&   \multicolumn{2}{c}{178.8 $\pm$ 104.5}&   \multicolumn{2}{c}{166.7 $\pm$ 97.6}&   \multicolumn{2}{c}{219.0 $\pm$ 88.8}\\
 &  4&   \multicolumn{2}{c}{629.7 $\pm$ 360.6}&   \multicolumn{2}{c}{702.0 $\pm$ 356.8}&   \multicolumn{2}{c}{752.3 $\pm$ 340.2}&   \multicolumn{2}{c}{750.3 $\pm$ 397.3}&   \multicolumn{2}{c}{756.2 $\pm$ 322.1}\\ \hline
 Freeway&  0
&   \multicolumn{2}{c}{21.6 $\pm$ 1.5}&   \multicolumn{2}{c}{21.7 $\pm$ 1.4}&   \multicolumn{2}{c}{21.9 $\pm$ 1.7}&   \multicolumn{2}{c}{22.4 $\pm$ 2.1}&   \multicolumn{2}{c}{21.2 $\pm$ 1.1}\\
 &  1
&   \multicolumn{2}{c}{9.6 $\pm$ 0.7}&   \multicolumn{2}{c}{9.5 $\pm$ 0.8}&   \multicolumn{2}{c}{9.7 $\pm$ 0.7}&   \multicolumn{2}{c}{9.5 $\pm$ 0.7}&   \multicolumn{2}{c}{9.6 $\pm$ 0.8}\\
 &  2
&   \multicolumn{2}{c}{13.9 $\pm$ 1.3}&   \multicolumn{2}{c}{13.7 $\pm$ 1.6}&   \multicolumn{2}{c}{13.9 $\pm$ 1.6}&   \multicolumn{2}{c}{13.9 $\pm$ 1.5}&   \multicolumn{2}{c}{14.1 $\pm$ 1.5}\\
 &  4&   \multicolumn{2}{c}{10.7 $\pm$ 0.8}&   \multicolumn{2}{c}{10.6 $\pm$ 0.9}&   \multicolumn{2}{c}{10.6 $\pm$ 0.6}&   \multicolumn{2}{c}{10.8 $\pm$ 0.8}&   \multicolumn{2}{c}{10.9 $\pm$ 0.7}\\ \hline
 Frostbite&  0
&   \multicolumn{2}{c}{125.3 $\pm$ 19.1}&   \multicolumn{2}{c}{1010.3 $\pm$ 480.0}&   \multicolumn{2}{c}{202.7 $\pm$ 5.1}&   \multicolumn{2}{c}{208.7 $\pm$ 3.4}&   \multicolumn{2}{c}{231.7 $\pm$ 246.8}\\
 &  1
&   \multicolumn{2}{c}{202.8 $\pm$ 44.2}&   \multicolumn{2}{c}{207.6 $\pm$ 49.1}&   \multicolumn{2}{c}{241.2 $\pm$ 24.3}&   \multicolumn{2}{c}{221.6 $\pm$ 38.2}&   \multicolumn{2}{c}{160.4 $\pm$ 56.1}\\
 &  2
&   \multicolumn{2}{c}{175.7 $\pm$ 37.3}&   \multicolumn{2}{c}{86.7 $\pm$ 29.9}&   \multicolumn{2}{c}{190.3 $\pm$ 34.9}&   \multicolumn{2}{c}{129.7 $\pm$ 50.1}&   \multicolumn{2}{c}{102.7 $\pm$ 48.7}\\
 &  4&   \multicolumn{2}{c}{1084.4 $\pm$ 891.6}&   \multicolumn{2}{c}{540.4 $\pm$ 403.9}&   \multicolumn{2}{c}{538.4 $\pm$ 487.6}&   \multicolumn{2}{c}{229.6 $\pm$ 14.1}&   \multicolumn{2}{c}{271.6 $\pm$ 26.2}\\
\hline
\end{tabular}
\end{table*}

\begin{table*}[h]
\caption{\textbf{Performance of  DQN variants (2/2)} with modified block conditions on 26 Atari tasks. The results show the optimal performing DQN variant --DQN, Double DQN, Noisy DQN, Dueling DQN, and Distributional DQN--  with  mean and standard deviation averaged thirty runs in each block conditions.}
\label{tb:2_2}
\centering
\begin{tabular}{l|c|@{\hspace{1mm}}llllllllll}
\hline
 Environments&  Blocks&  \multicolumn{2}{c}{DQN}&  \multicolumn{2}{c}{Double DQN}&  \multicolumn{2}{c}{Noisy DQN}&  \multicolumn{2}{c}{Dueling DQN}&  \multicolumn{2}{c}{Distributional DQN}\\ \hline
 Gopher&  0
&   \multicolumn{2}{c}{124.0 $\pm$ 134.6}&   \multicolumn{2}{c}{170.7 $\pm$ 56.0}&   \multicolumn{2}{c}{203.3 $\pm$ 130.3}&   \multicolumn{2}{c}{262.0 $\pm$ 203.2}&   \multicolumn{2}{c}{216.0 $\pm$ 193.0}\\
 &  1
&   \multicolumn{2}{c}{224.0 $\pm$ 160.8}&   \multicolumn{2}{c}{221.0 $\pm$ 193.0}&   \multicolumn{2}{c}{159.0 $\pm$ 103.9}&   \multicolumn{2}{c}{373.7 $\pm$ 184.0}&   \multicolumn{2}{c}{183.3 $\pm$ 138.7}\\
 &  2
&   \multicolumn{2}{c}{467.3 $\pm$ 245.5}&   \multicolumn{2}{c}{311.3 $\pm$ 197.2}&   \multicolumn{2}{c}{361.3 $\pm$ 175.6}&   \multicolumn{2}{c}{357.3 $\pm$ 228.5}&   \multicolumn{2}{c}{401.3 $\pm$ 261.4}\\
 &  4&   \multicolumn{2}{c}{284.3 $\pm$ 193.1}&   \multicolumn{2}{c}{104.0 $\pm$ 124.4}&   \multicolumn{2}{c}{110.7 $\pm$ 155.1}&   \multicolumn{2}{c}{267.3 $\pm$ 267.9}&   \multicolumn{2}{c}{126.7 $\pm$ 145.7}\\ \hline
 Hero&  0
&   \multicolumn{2}{c}{341.5 $\pm$ 754.5}&   \multicolumn{2}{c}{1635.8 $\pm$ 1078.9}&   \multicolumn{2}{c}{1795.7 $\pm$ 320.6}&   \multicolumn{2}{c}{2930.0 $\pm$ 72.3}&   \multicolumn{2}{c}{2453.3 $\pm$ 41.8}\\
 &  1
&   \multicolumn{2}{c}{2354.3 $\pm$ 899.7}&   \multicolumn{2}{c}{2252.0 $\pm$ 1093.1}&   \multicolumn{2}{c}{932.2 $\pm$ 1220.6}&   \multicolumn{2}{c}{2898.7 $\pm$ 116.4}&   \multicolumn{2}{c}{2332.2 $\pm$ 1119.4}\\
 &  2
&   \multicolumn{2}{c}{2400.7 $\pm$ 1053.3}&   \multicolumn{2}{c}{1171.2 $\pm$ 1189.1}&   \multicolumn{2}{c}{1533.2 $\pm$ 854.2}&   \multicolumn{2}{c}{2996.8 $\pm$ 93.7}&   \multicolumn{2}{c}{2603.8 $\pm$ 993.8}\\
 &  4&   \multicolumn{2}{c}{2547.2 $\pm$ 456.6}&   \multicolumn{2}{c}{2629.8 $\pm$ 681.1}&   \multicolumn{2}{c}{1451.8 $\pm$ 1351.8}&   \multicolumn{2}{c}{2454.5 $\pm$ 174.0}&   \multicolumn{2}{c}{2662.0 $\pm$ 27.6}\\ \hline
 Jamesbond&  0
&   \multicolumn{2}{c}{5.0 $\pm$ 15.0}&   \multicolumn{2}{c}{83.3 $\pm$ 69.9}&   \multicolumn{2}{c}{168.3 $\pm$ 89.0}&   \multicolumn{2}{c}{73.3 $\pm$ 47.8}&   \multicolumn{2}{c}{31.7 $\pm$ 32.9}\\
 &  1
&   \multicolumn{2}{c}{72.0 $\pm$ 48.7}&   \multicolumn{2}{c}{47.0 $\pm$ 36.9}&   \multicolumn{2}{c}{27.0 $\pm$ 24.9}&   \multicolumn{2}{c}{49.0 $\pm$ 33.3}&   \multicolumn{2}{c}{40.0 $\pm$ 34.9}\\
 &  2
&   \multicolumn{2}{c}{111.7 $\pm$ 52.7}&   \multicolumn{2}{c}{73.3 $\pm$ 60.2}&   \multicolumn{2}{c}{71.7 $\pm$ 49.5}&   \multicolumn{2}{c}{56.7 $\pm$ 44.2}&   \multicolumn{2}{c}{73.3 $\pm$ 46.1}\\
 &  4&   \multicolumn{2}{c}{46.0 $\pm$ 47.6}&   \multicolumn{2}{c}{51.0 $\pm$ 43.2}&   \multicolumn{2}{c}{25.0 $\pm$ 25.8}&   \multicolumn{2}{c}{16.0 $\pm$ 15.0}&   \multicolumn{2}{c}{76.0 $\pm$ 38.5}\\ \hline
 Kangaroo&  0
&   \multicolumn{2}{c}{446.7 $\pm$ 160.7}&   \multicolumn{2}{c}{300.0 $\pm$ 100.0}&   \multicolumn{2}{c}{346.7 $\pm$ 88.4}&   \multicolumn{2}{c}{426.7 $\pm$ 68.0}&   \multicolumn{2}{c}{600.0 $\pm$ 0.0}\\
 &  1
&   \multicolumn{2}{c}{666.7 $\pm$ 174.8}&   \multicolumn{2}{c}{641.7 $\pm$ 271.4}&   \multicolumn{2}{c}{508.3 $\pm$ 120.5}&   \multicolumn{2}{c}{708.3 $\pm$ 289.3}&   \multicolumn{2}{c}{791.7 $\pm$ 329.7}\\
 &  2
&   \multicolumn{2}{c}{293.3 $\pm$ 235.1}&   \multicolumn{2}{c}{466.7 $\pm$ 130.0}&   \multicolumn{2}{c}{73.3 $\pm$ 109.3}&   \multicolumn{2}{c}{546.7 $\pm$ 145.4}&   \multicolumn{2}{c}{373.3 $\pm$ 276.8}\\
 &  4&   \multicolumn{2}{c}{308.3 $\pm$ 123.9}&   \multicolumn{2}{c}{250.0 $\pm$ 0.0}&   \multicolumn{2}{c}{0.0 $\pm$ 0.0}&   \multicolumn{2}{c}{250.0 $\pm$ 0.0}&   \multicolumn{2}{c}{500.0 $\pm$ 0.0}\\ \hline
 Krull&  0
&   \multicolumn{2}{c}{1912.0 $\pm$ 516.7}&   \multicolumn{2}{c}{2859.7 $\pm$ 959.1}&   \multicolumn{2}{c}{5181.0 $\pm$ 4774.0}&   \multicolumn{2}{c}{2008.3 $\pm$ 428.6}&   \multicolumn{2}{c}{1804.7 $\pm$ 1122.8}\\
 &  1
&   \multicolumn{2}{c}{2343.2 $\pm$ 1443.0}&   \multicolumn{2}{c}{1477.3 $\pm$ 402.5}&   \multicolumn{2}{c}{2460.5 $\pm$ 1771.8}&   \multicolumn{2}{c}{1608.0 $\pm$ 495.2}&   \multicolumn{2}{c}{1109.5 $\pm$ 625.9}\\
 &  2
&   \multicolumn{2}{c}{3833.7 $\pm$ 1486.3}&   \multicolumn{2}{c}{3425.3 $\pm$ 602.1}&   \multicolumn{2}{c}{5286.7 $\pm$ 4426.4}&   \multicolumn{2}{c}{3136.0 $\pm$ 676.6}&   \multicolumn{2}{c}{3354.3 $\pm$ 489.8}\\
 &  4&   \multicolumn{2}{c}{1206.8 $\pm$ 295.8}&   \multicolumn{2}{c}{857.8 $\pm$ 381.6}&   \multicolumn{2}{c}{1286.8 $\pm$ 632.7}&   \multicolumn{2}{c}{1004.2 $\pm$ 184.0}&   \multicolumn{2}{c}{990.5 $\pm$ 241.6}\\ \hline
 KungFuMaster&  0
&   \multicolumn{2}{c}{0.0 $\pm$ 0.0}&   \multicolumn{2}{c}{0.0 $\pm$ 0.0}&   \multicolumn{2}{c}{0.0 $\pm$ 0.0}&   \multicolumn{2}{c}{0.0 $\pm$ 0.0}&   \multicolumn{2}{c}{0.0 $\pm$ 0.0}\\
 &  1
&   \multicolumn{2}{c}{1173.3 $\pm$ 620.2}&   \multicolumn{2}{c}{1160.0 $\pm$ 748.2}&   \multicolumn{2}{c}{796.7 $\pm$ 576.5}&   \multicolumn{2}{c}{1070.0 $\pm$ 716.5}&   \multicolumn{2}{c}{1320.0 $\pm$ 735.5}\\
 &  2
&   \multicolumn{2}{c}{1570.0 $\pm$ 725.3}&   \multicolumn{2}{c}{1366.7 $\pm$ 785.8}&   \multicolumn{2}{c}{1176.7 $\pm$ 633.3}&   \multicolumn{2}{c}{1643.3 $\pm$ 755.7}&   \multicolumn{2}{c}{1250.0 $\pm$ 557.8}\\
 &  4&   \multicolumn{2}{c}{0.0 $\pm$ 0.0}&   \multicolumn{2}{c}{0.0 $\pm$ 0.0}&   \multicolumn{2}{c}{0.0 $\pm$ 0.0}&   \multicolumn{2}{c}{0.0 $\pm$ 0.0}&   \multicolumn{2}{c}{0.0 $\pm$ 0.0}\\ \hline
 MsPacman&  0
&   \multicolumn{2}{c}{1030.3 $\pm$ 187.5}&   \multicolumn{2}{c}{774.3 $\pm$ 457.2}&   \multicolumn{2}{c}{1197.3 $\pm$ 1065.6}&   \multicolumn{2}{c}{1360.3 $\pm$ 625.5}&   \multicolumn{2}{c}{1151.7 $\pm$ 329.8}\\
 &  1
&   \multicolumn{2}{c}{504.3 $\pm$ 277.3}&   \multicolumn{2}{c}{524.0 $\pm$ 329.2}&   \multicolumn{2}{c}{681.7 $\pm$ 338.6}&   \multicolumn{2}{c}{591.7 $\pm$ 370.4}&   \multicolumn{2}{c}{304.8 $\pm$ 206.3}\\
 &  2
&   \multicolumn{2}{c}{503.3 $\pm$ 240.8}&   \multicolumn{2}{c}{481.7 $\pm$ 260.3}&   \multicolumn{2}{c}{649.3 $\pm$ 269.1}&   \multicolumn{2}{c}{496.7 $\pm$ 178.3}&   \multicolumn{2}{c}{548.0 $\pm$ 270.3}\\
 &  4&   \multicolumn{2}{c}{587.0 $\pm$ 413.1}&   \multicolumn{2}{c}{1036.8 $\pm$ 402.6}&   \multicolumn{2}{c}{1190.7 $\pm$ 548.9}&   \multicolumn{2}{c}{1185.3 $\pm$ 686.0}&   \multicolumn{2}{c}{728.5 $\pm$ 202.0}\\ \hline
 Pong&  0
&   \multicolumn{2}{c}{-20.4 $\pm$ 0.5}&   \multicolumn{2}{c}{-17.9 $\pm$ 4.8}&   \multicolumn{2}{c}{-20.5 $\pm$ 0.8}&   \multicolumn{2}{c}{-21.0 $\pm$ 0.0}&   \multicolumn{2}{c}{-21.0 $\pm$ 0.0}\\
 &  1
&   \multicolumn{2}{c}{-19.8 $\pm$ 1.1}&   \multicolumn{2}{c}{-20.7 $\pm$ 0.6}&   \multicolumn{2}{c}{-20.7 $\pm$ 0.4}&   \multicolumn{2}{c}{-20.5 $\pm$ 0.6}&   \multicolumn{2}{c}{-20.8 $\pm$ 0.3}\\
 &  2
&   \multicolumn{2}{c}{-19.6 $\pm$ 1.7}&   \multicolumn{2}{c}{-19.4 $\pm$ 1.3}&   \multicolumn{2}{c}{-20.5 $\pm$ 0.8}&   \multicolumn{2}{c}{-19.8 $\pm$ 1.4}&   \multicolumn{2}{c}{-20.5 $\pm$ 0.6}\\
 &  4&   \multicolumn{2}{c}{-20.5 $\pm$ 0.9}&   \multicolumn{2}{c}{-20.9 $\pm$ 0.2}&   \multicolumn{2}{c}{-20.8 $\pm$ 0.2}&   \multicolumn{2}{c}{-20.9 $\pm$ 0.2}&   \multicolumn{2}{c}{-21.0 $\pm$ 0.0}\\ \hline
 PrivateEye&  0
&   \multicolumn{2}{c}{100.0 $\pm$ 0.0}&   \multicolumn{2}{c}{93.3 $\pm$ 24.9}&   \multicolumn{2}{c}{100.0 $\pm$ 0.0}&   \multicolumn{2}{c}{96.7 $\pm$ 18.0}&   \multicolumn{2}{c}{0.0 $\pm$ 0.0}\\
 &  1
&   \multicolumn{2}{c}{4465.0 $\pm$ 289.9}&   \multicolumn{2}{c}{1788.6 $\pm$ 4094.8}&   \multicolumn{2}{c}{100.0 $\pm$ 0.0}&   \multicolumn{2}{c}{3907.9 $\pm$ 1891.2}&   \multicolumn{2}{c}{4683.6 $\pm$ 1253.0}\\
 &  2
&   \multicolumn{2}{c}{2088.0 $\pm$ 4537.9}&   \multicolumn{2}{c}{1834.3 $\pm$ 2431.9}&   \multicolumn{2}{c}{96.7 $\pm$ 18.0}&   \multicolumn{2}{c}{3339.9 $\pm$ 1884.0}&   \multicolumn{2}{c}{4173.3 $\pm$ 218.4}\\
 &  4&   \multicolumn{2}{c}{100.0 $\pm$ 0.0}&   \multicolumn{2}{c}{70.0 $\pm$ 45.8}&   \multicolumn{2}{c}{46.7 $\pm$ 49.9}&   \multicolumn{2}{c}{4702.7 $\pm$ 1520.7}&   \multicolumn{2}{c}{4017.5 $\pm$ 1959.4}\\ \hline
 Qbert&  0
&   \multicolumn{2}{c}{657.5 $\pm$ 94.2}&   \multicolumn{2}{c}{357.5 $\pm$ 241.6}&   \multicolumn{2}{c}{175.0 $\pm$ 0.0}&   \multicolumn{2}{c}{675.0 $\pm$ 0.0}&   \multicolumn{2}{c}{415.0 $\pm$ 53.9}\\
 &  1
&   \multicolumn{2}{c}{173.3 $\pm$ 81.2}&   \multicolumn{2}{c}{257.7 $\pm$ 141.2}&   \multicolumn{2}{c}{123.0 $\pm$ 39.8}&   \multicolumn{2}{c}{299.5 $\pm$ 159.4}&   \multicolumn{2}{c}{168.2 $\pm$ 89.4}\\
 &  2
&   \multicolumn{2}{c}{242.5 $\pm$ 110.3}&   \multicolumn{2}{c}{375.8 $\pm$ 229.7}&   \multicolumn{2}{c}{406.7 $\pm$ 269.7}&   \multicolumn{2}{c}{348.3 $\pm$ 198.7}&   \multicolumn{2}{c}{332.5 $\pm$ 186.4}\\
 &  4&   \multicolumn{2}{c}{872.7 $\pm$ 894.5}&   \multicolumn{2}{c}{297.5 $\pm$ 403.8}&   \multicolumn{2}{c}{690.0 $\pm$ 489.9}&   \multicolumn{2}{c}{570.7 $\pm$ 499.8}&   \multicolumn{2}{c}{975.0 $\pm$ 300.0}\\ \hline
 RoadRunner&  0
&   \multicolumn{2}{c}{2520.0 $\pm$ 1265.8}&   \multicolumn{2}{c}{4036.7 $\pm$ 1677.4}&   \multicolumn{2}{c}{4700.0 $\pm$ 2134.5}&   \multicolumn{2}{c}{1546.7 $\pm$ 623.8}&   \multicolumn{2}{c}{2256.7 $\pm$ 1480.0}\\
 &  1
&   \multicolumn{2}{c}{1273.3 $\pm$ 673.5}&   \multicolumn{2}{c}{1128.3 $\pm$ 510.0}&   \multicolumn{2}{c}{1108.3 $\pm$ 776.4}&   \multicolumn{2}{c}{1335.0 $\pm$ 562.9}&   \multicolumn{2}{c}{1291.7 $\pm$ 477.5}\\
 &  2
&   \multicolumn{2}{c}{1023.3 $\pm$ 899.5}&   \multicolumn{2}{c}{1140.0 $\pm$ 927.9}&   \multicolumn{2}{c}{1246.7 $\pm$ 797.8}&   \multicolumn{2}{c}{1456.7 $\pm$ 891.7}&   \multicolumn{2}{c}{1346.7 $\pm$ 777.5}\\
 &  4&   \multicolumn{2}{c}{2875.0 $\pm$ 1705.2}&   \multicolumn{2}{c}{2931.7 $\pm$ 1439.6}&   \multicolumn{2}{c}{2940.0 $\pm$ 1089.1}&   \multicolumn{2}{c}{1406.7 $\pm$ 373.2}&   \multicolumn{2}{c}{2183.3 $\pm$ 879.3}\\ \hline
 Seaquest&  0
&   \multicolumn{2}{c}{239.3 $\pm$ 54.5}&   \multicolumn{2}{c}{234.0 $\pm$ 80.4}&   \multicolumn{2}{c}{126.7 $\pm$ 33.2}&   \multicolumn{2}{c}{218.7 $\pm$ 73.2}&   \multicolumn{2}{c}{246.0 $\pm$ 64.1}\\
 &  1
&   \multicolumn{2}{c}{428.3 $\pm$ 123.6}&   \multicolumn{2}{c}{599.3 $\pm$ 243.0}&   \multicolumn{2}{c}{679.7 $\pm$ 253.0}&   \multicolumn{2}{c}{618.0 $\pm$ 191.3}&   \multicolumn{2}{c}{459.7 $\pm$ 122.5}\\
 &  2
&   \multicolumn{2}{c}{236.0 $\pm$ 77.4}&   \multicolumn{2}{c}{227.3 $\pm$ 85.2}&   \multicolumn{2}{c}{280.7 $\pm$ 87.3}&   \multicolumn{2}{c}{212.7 $\pm$ 72.6}&   \multicolumn{2}{c}{318.7 $\pm$ 89.1}\\
 &  4&   \multicolumn{2}{c}{573.3 $\pm$ 258.8}&   \multicolumn{2}{c}{515.0 $\pm$ 84.8}&   \multicolumn{2}{c}{441.7 $\pm$ 140.9}&   \multicolumn{2}{c}{570.0 $\pm$ 110.8}&   \multicolumn{2}{c}{563.3 $\pm$ 136.6}\\ \hline
 UpNDown&  0
&   \multicolumn{2}{c}{1948.3 $\pm$ 875.5}&   \multicolumn{2}{c}{1892.0 $\pm$ 315.7}&   \multicolumn{2}{c}{2248.7 $\pm$ 750.4}&   \multicolumn{2}{c}{1699.3 $\pm$ 465.6}&   \multicolumn{2}{c}{2940.3 $\pm$ 1104.8}\\
 &  1
&   \multicolumn{2}{c}{1471.3 $\pm$ 1151.0}&   \multicolumn{2}{c}{1669.7 $\pm$ 913.0}&   \multicolumn{2}{c}{1353.3 $\pm$ 1116.6}&   \multicolumn{2}{c}{1518.7 $\pm$ 980.0}&   \multicolumn{2}{c}{1821.0 $\pm$ 1323.2}\\
 &  2
&   \multicolumn{2}{c}{1708.3 $\pm$ 808.9}&   \multicolumn{2}{c}{1902.3 $\pm$ 1026.9}&   \multicolumn{2}{c}{1713.3 $\pm$ 805.0}&   \multicolumn{2}{c}{1933.3 $\pm$ 994.2}&   \multicolumn{2}{c}{1674.3 $\pm$ 869.9}\\
 &  4&   \multicolumn{2}{c}{925.3 $\pm$ 567.8}&   \multicolumn{2}{c}{2063.7 $\pm$ 984.9}&   \multicolumn{2}{c}{1760.7 $\pm$ 884.7}&   \multicolumn{2}{c}{1913.7 $\pm$ 652.5}&   \multicolumn{2}{c}{3297.7 $\pm$ 1340.7}\\
\hline
\end{tabular}
\end{table*}
\bibliography{mybibfile}